\documentclass[10pt,twocolumn,letterpaper]{article}

\usepackage{cvpr}              

\usepackage[accsupp]{axessibility}
\usepackage{graphicx}
\usepackage{amsmath}
\usepackage{amssymb}
\usepackage{booktabs}
\usepackage{enumitem}
\usepackage{multirow}
\usepackage{xcolor}
\usepackage{bbding}
\newtheorem{definition}{Definition}
\newtheorem{proposition}{Proposition}
\newcommand{\tablestyle}[2]{\setlength{\tabcolsep}{#1}\renewcommand{\arraystretch}{#2}\centering\small}
\usepackage{makecell}

\usepackage{color} 

\usepackage{multicol}

\usepackage[pagebackref,breaklinks,colorlinks]{hyperref}

\usepackage[capitalize]{cleveref}
\crefname{section}{Sec.}{Secs.}
\Crefname{section}{Section}{Sections}
\Crefname{table}{Table}{Tables}
\crefname{table}{Tab.}{Tabs.}

\definecolor{themeblue}{RGB}{57, 162, 219}

\def\cvprPaperID{894} 
\def\confName{CVPR}
\def\confYear{2023}

\begin{document}

\title{Enlarging Instance-specific and Class-specific Information for \\ Open-set Action Recognition}

\author{Jun Cen$^{1,2}$\thanks{Work done as an intern at Alibaba DAMO Academy.} \quad Shiwei Zhang$^{2}$ \quad Xiang Wang$^{3}$ \quad Yixuan Pei$^{4}$ \\ Zhiwu Qing$^{3}$ \quad
Yingya Zhang$^{2}$ \quad Qifeng Chen$^{1}$\\
\textsuperscript{1}The Hong Kong University of Science and Technology \quad
\textsuperscript{2}Alibaba Group \\
\textsuperscript{3}Huazhong University of Science and Technology \quad
\textsuperscript{4}Xi'an Jiaotong University\\
\tt\small{jcenaa@connect.ust.hk,}
\tt\small{\{zhangjin.zsw,yingya.zyy\}@alibaba-inc.com,}\\
\tt\small{\{wxiang,qzw\}@hust.edu.cn,}
\tt\small{peiyixuan@stu.xjtu.edu.cn,}
\tt\small{cqf@ust.hk}}


\maketitle

\def\eg{\emph{e.g}\onedot} \def\Eg{\emph{E.g}\onedot}
\def\ie{\emph{i.e}\onedot} \def\Ie{\emph{I.e}\onedot}
\def\cf{\emph{cf}\onedot} \def\Cf{\emph{Cf}\onedot}
\def\etc{\emph{etc}\onedot} \def\vs{\emph{vs}\onedot}
\def\wrt{w.r.t\onedot} \def\dof{d.o.f\onedot}
\def\etal{\emph{et al}\onedot}
\makeatother

\newcommand{\va}{\mathbf{a}}
\newcommand{\vb}{\mathbf{b}}
\newcommand{\vc}{\mathbf{c}}
\newcommand{\vd}{\mathbf{d}}
\newcommand{\ve}{\mathbf{e}}
\newcommand{\vf}{\mathbf{f}}
\newcommand{\vg}{\mathbf{g}}
\newcommand{\vh}{\mathbf{h}}
\newcommand{\vi}{\mathbf{i}}
\newcommand{\vj}{\mathbf{j}}
\newcommand{\vk}{\mathbf{k}}
\newcommand{\vl}{\mathbf{l}}
\newcommand{\vm}{\mathbf{m}}
\newcommand{\vn}{\mathbf{n}}
\newcommand{\vo}{\mathbf{o}}
\newcommand{\vp}{\mathbf{p}}
\newcommand{\vq}{\mathbf{q}}
\newcommand{\vr}{\mathbf{r}}
\newcommand{\vt}{\mathbf{t}}
\newcommand{\vu}{\mathbf{u}}
\newcommand{\vv}{\mathbf{v}}
\newcommand{\vw}{\mathbf{w}}
\newcommand{\vx}{\mathbf{x}}
\newcommand{\vy}{\mathbf{y}}
\newcommand{\vz}{\mathbf{z}}
\newcommand{\vT}{\mathbf{T}}
\newcommand{\vW}{\mathbf{W}}
\newcommand{\vX}{\mathbf{X}}
\newcommand{\vY}{\mathbf{Y}}
\newcommand{\vZ}{\mathbf{Z}}

\begin{abstract}
Open-set action recognition is to reject unknown human action cases which are out of the distribution of the training set. 
Existing methods mainly focus on learning better uncertainty scores but dismiss the importance of feature representations.
We find that features with richer semantic diversity can significantly improve the open-set performance under the same uncertainty scores.
In this paper, we begin with analyzing the feature representation behavior in the open-set action recognition (OSAR) problem based on the information bottleneck (IB) theory, and propose to enlarge the instance-specific (IS) and class-specific (CS) information contained in the feature for better performance.
To this end, a novel Prototypical Similarity Learning (PSL) framework is proposed to keep the instance variance within the same class to retain more IS information.
Besides, we notice that unknown samples sharing similar appearances to known samples are easily misclassified as known classes.
To alleviate this issue, video shuffling is further introduced in our PSL to learn distinct temporal information between original and shuffled samples, which we find enlarges the CS information.
Extensive experiments demonstrate that the proposed PSL can significantly boost both the open-set and closed-set performance and achieves state-of-the-art results on multiple benchmarks. Code is available at \url{https://github.com/Jun-CEN/PSL}.

\end{abstract}

\section{Introduction}
\label{sec:intro}

Deep learning methods for video action recognition have developed very fast and achieved remarkable performance in recent years~\cite{lin2019tsm, feichtenhofer2019slowfast, i3d, yang2020temporal}. 
\begin{figure}[t]
    \centering
    \includegraphics[width=0.99\linewidth]{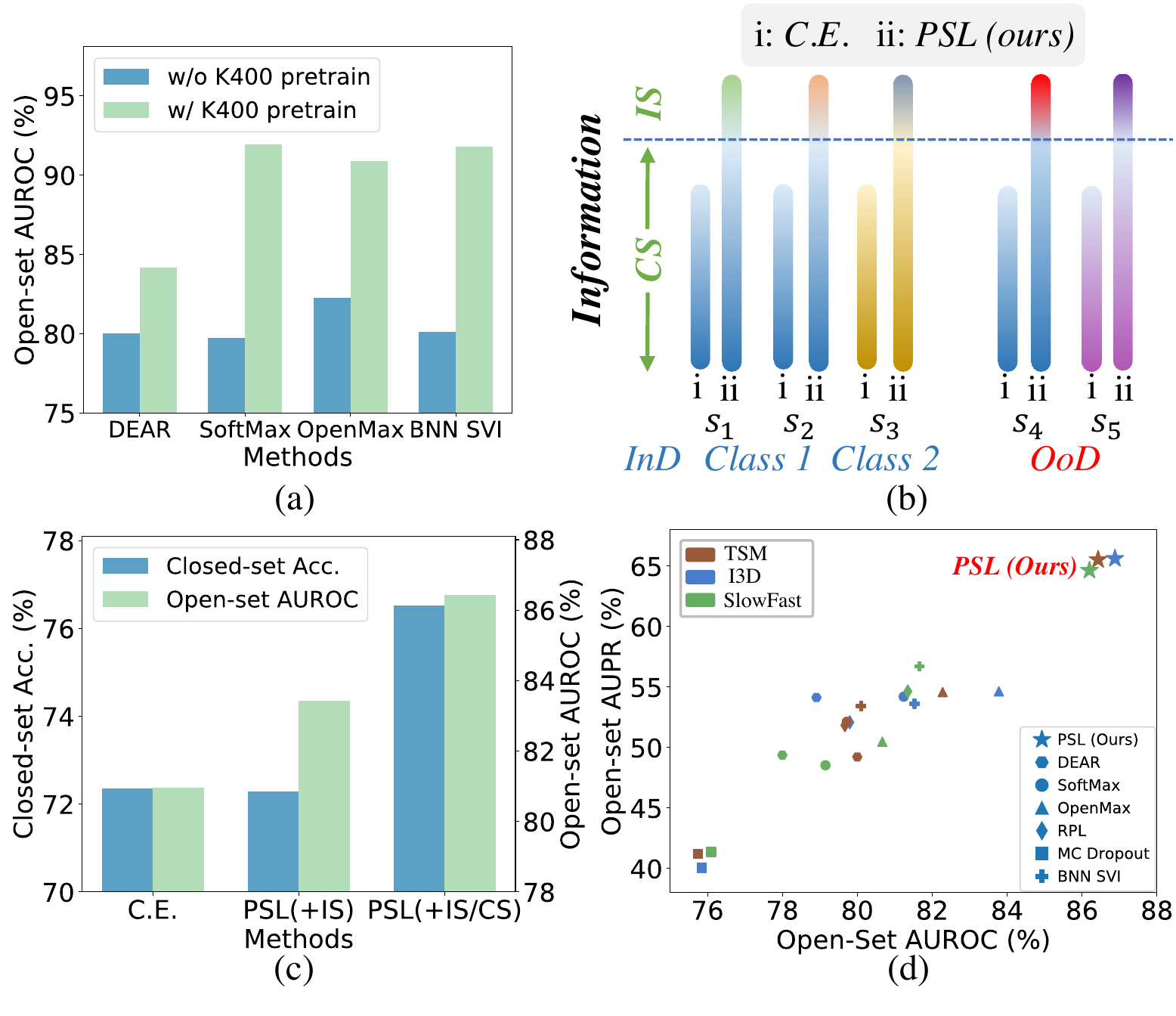}
    \vspace{-0.5cm}
    \caption{(a) Richer semantic features brought by the pretraining can significantly improve the open-set performance. (b) Information in the feature is divided into IS and CS information. $s_4$ can be identified as OoD since it has distinct IS information (IS bars in different colors) with $s_1$ and $s_2$, while $s_5$ has distinct CS information (CS bars in different colors) with all InD samples so it may be OoD. Our PSL aims to learn more IS and CS information (bars in longer lengths) than Cross-Entropy (C.E.). (c) Both enlarged IS and CS information boosts the open-set performance. (d) Our PSL achieves the best OSAR performance.}
    \label{fig:1}
\end{figure}
However, these methods operate under the \emph{closed-set} condition, \textit{i.e.}, to classify all videos into one of the classes encountered during training. This closed-set condition is not practical in the real-world scenario, as videos whose classes are beyond the range of the training set will be misclassified as one of the known classes. 
Therefore, \emph{open-set action recognition} (OSAR) is proposed to require the network to correctly classify in-distribution (InD) samples and identify out-of-distribution (OoD) samples. InD and OoD classes refer to classes involved and not involved in the training set, respectively.

Open-set video action recognition is systematically studied in the recent work~\cite{bao2021evidential}, in which they transfer the existing methods for open-set image recognition into the video domain~\cite{hendrycks2016baseline,gal2016dropout,bendale2016towards,kendall2017uncertainties} as the baselines, and propose their own method to introduce deep evidential learning~\cite{amini2020deep} to calculate the uncertainty and propose a contrastive evidential debiasing module to alleviate the appearance bias issue in the video domain. All of these methods tend to improve the OSAR performance by calculating a better uncertainty score, based on the feature representations extracted by the neural network (NN). However, the main purpose of training in these methods is still to classify InD samples, which determines the learned feature representations are merely sufficient for InD classification. We find that almost all methods have a significantly better open-set performance when the NN is pretrained with a large dataset (\cref{fig:1} (a)), so we argue that the diversity of feature representation is extremely important for the OSAR task. Therefore, we propose to boost the open-set ability from the feature representation perspective rather than finding a better uncertainty score.

We first analyze the feature representation behavior in the open-set problem based on the information bottleneck (IB) theory~\cite{tishby2015deep,wang2022rethinking}. We divide the information of the feature into \emph{Instance-Specific (IS)}
and \emph{Class-Specific (CS) information}. CS information is used for inter-class recognition, so it is similar for samples within the same class but different for samples from other classes. IS information is the special information of each sample within the same class, as two samples cannot be exactly the same even if they belong to the same class. Both CS and IS information are crucial for the open-set task, as illustrated in \cref{fig:1} (b), where $s_4$ and $s_5$ can be identified as OoD samples based on the IS and CS information, respectively. We find that the closed-set classification setting tends to eliminate IS information during training, and cannot fully extract the minimum sufficient CS information for the classification task, so we aim to enlarge IS and CS information in learned feature representations for better OSAR performance.

To enlarge the IS information, we propose the \emph{Prototypical Similarity Learning} (PSL) framework, in which the representation of an instance is encouraged to have less than 1 similarity with the corresponding prototype. In this way, we encourage the IS information to be retained and not eliminated. In addition, \cite{bao2021evidential} finds that OoD videos can be easily classified as InD videos in a similar appearance. To alleviate this issue, we introduce the shuffled video into PSL and make it have less than 1 similarity with the original sample. As the shuffled video almost shares the same appearance information with the original one, we encourage the similarity to be less than 1 so that the network can extract the distinct temporal information among them. We find this technique actually enlarges the CS information in the feature representation. \cref{fig:1} (c) shows that enlarging the IS information is helpful for the open-set performance, and more CS information can further benefit the open-set and closed-set performance. 
To summarize, our contributions include:
\begin{itemize}[leftmargin=*, itemsep=0 pt, topsep=0 pt, parsep=0 pt]
    \item We provide a novel perspective to analyze the open-set recognition task based on the information bottleneck theory, and find that the classical closed-set cross-entropy tends to eliminate the IS information which is helpful to identify OoD samples.
    \item We propose to enlarge the IS and CS information for better OSAR performance. Specifically, PSL is designed to retain the IS information in the features, and we involve video shuffling in PSL to learn more CS information. 
    \item Experiments on multiple datasets and backbones show our PSL’s superiority over a large margin compared to other state-of-the-art counterparts, as shown in \cref{fig:1} (d).
\end{itemize}
\section{Related Work}
\label{sec:related}

\noindent \textbf{Action Recognition.}
Most recent approaches for action recognition are to exploit appearance and motion cues jointly and achieve remarkable success~\cite{feichtenhofer2019slowfast,i3d,lin2019tsm,huang2021tada,qing2022learning,wang2021oadtr,pei2022learning}.
Typically, two-stream networks~\cite{two-stream,two-stream-2,TSN} consist of two branches that explore spatial information and temporal dynamics, respectively.
%
Some attempts~\cite{lin2019tsm,r2+1d,TDN} introduce additional temporal mining operations to overcome the limited temporal information extraction ability of 2D CNN.
3D CNN-based methods~\cite{feichtenhofer2019slowfast, i3d,C3D} inflated 2D kernels for joint spatio-temporal modeling.
\cite{bai2020prototype} proposes the prototype similarity learning which pushes the learned representation to the corresponding prototype as close as possible, while our PSL keeps the differences among the same class.

\noindent \textbf{Open-set Action Recognition.} The related work of OSAR is limited~\cite{krishnan2018bar,shu2018odn,yang2019open,bao2021evidential}. Recently, \cite{bao2021evidential} systematically studies the OSAR problem and transfers several open-set image recognition methods to the video domain, including SoftMax~\cite{hendrycks2016baseline}, MC Dropout~\cite{gal2016dropout}, OpenMax~\cite{bendale2016towards}, and RPL~\cite{chen2020learning}. In the benchmark of~\cite{bao2021evidential}, the only two methods designed specifically for the video domain are BNN SVI~\cite{krishnan2018bar} and their proposed DEAR. BNN SVI is a Bayesian NN application in the OSAR, while DEAR adopts the deep evidential learning~\cite{amini2020deep} to calculate the uncertainty, and utilizes two modules to alleviate the over-confidence prediction and appearance bias problem, respectively. Existing methods pursue better uncertainty scores, while the objective of our PSL is to learn more diverse feature representations for better open-set distinguishability.

\noindent \textbf{Information Bottleneck Theory.} Based on the IB theory~\cite{tishby2000information,tishby2015deep}, the NN intends to extract minimum sufficient information of the inputs for the current task. More recent~\cite{tian2020makes,federici2020learning,wang2022rethinking} adopt the IB theory on unsupervised contrastive learning to analyze the representation learning behavior under the corresponding tasks. In this work, we provide a new view to analyze the OSAR problem based on the IB theory.
\section{Information Analysis in OSAR}
\label{sec:ana}

\subsection{Prototypical Learning}
Let $f$ be the encoder to extract the information for an input video sample $x$ and output the feature representation $z = f(x), z \in \mathbb{R}^d$. We first define a prototypical learning (PL) loss~\cite{yang2018robust}, which is a general version of the cross-entropy (C.E.) loss:
\begin{equation}
\vspace{-0.2cm}
    \mathcal{L}_{PL} = - \log \frac{\exp (\frac{z^T k_{i}} {\tau}) }{ \exp (\frac{z^T k_{i}}{ \tau}) + \sum\limits_{n \in K_i^{-}} \exp (\frac{z^T  n}{ \tau})},
    \label{eq:PL}
    \vspace{-0.2cm}
\end{equation}
where $i$ is the ground truth label of $x$, $k_i \in \mathbb{R}^d$ is the prototype for class $i$, $\tau$ is a temperature parameter, $K_i^{-}=\left \{ k_j| j \in \left \{ 1,2,...,N \right \}, j\ne i\right \}$ is the negative prototype set, and $N$ is the number of InD classes. Note that $z$ and $k_i$ are normalized by L2 norm, so that $z^T k_{i}$ is the cosine similarity. If we regard prototypes as the row vector of the linear classifier $W \in \mathbb{R}^{N \times d}$, and do not normalize $z$ and $k$ as well as remove $\tau$, $\mathcal{L}_{PL}$ will degenerate to the C.E. loss. We introduce the $\mathcal{L}_{PL}$ so that we can directly manipulate the feature representation $z$.

\subsection{Information Analysis of OSAR}
\begin{figure}[t]
    \centering
\includegraphics[width=0.99\linewidth]{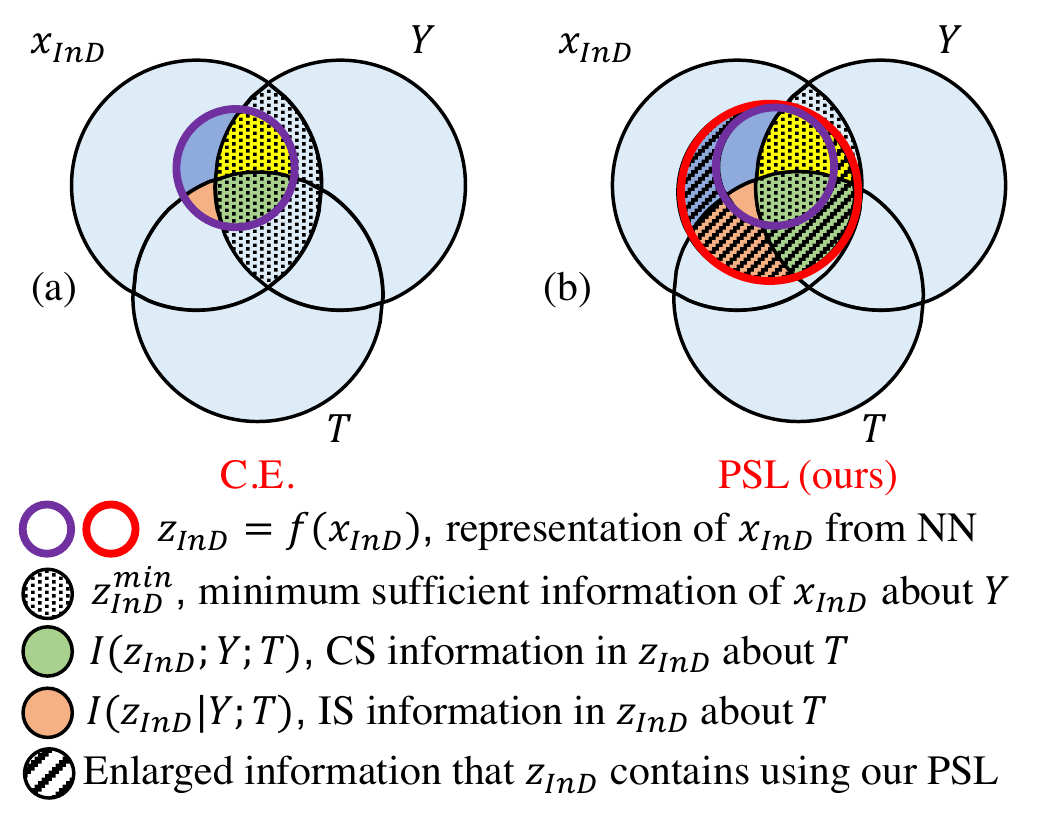}
    \vspace{-0.4cm}
    \caption{The neural network (NN) can only extract limited representations $z_{InD}$ of the InD sample $x_{InD}$ for the current task $Y$ (predict the closed-set label), which is not diverse enough for the task $T$ (distinguish OoD samples), as green and orange areas are small in (a). In our PSL, we encourage the NN to learn a more diverse representation so that more IS and CS information about $T$ are contained.}
    \label{fig:ana}
\end{figure}

Let $x_{InD}, z_{InD}$, and $Y$ be the random variables of InD sample, extracted representation of InD sample, and the task to predict the label of $x_{InD}$, where $z_{InD}=f(x_{InD})$. Given the joint distribution of $p(x_{InD},Y)$, the relevant information between $x_{InD}$ and $Y$ is defined as $I(x_{InD},Y)$, where $I$ denotes the mutual information~\cite{tishby2000information}. The learned representation $z_{InD}$ satisfies:
\vspace{-0.2cm}
\begin{equation}
    I(x_{InD}; z_{InD}) = \underbrace{I(x_{InD}; z_{InD}|Y)}_{IS} + \underbrace{I(z_{InD};Y)}_{CS},
    \label{eq:sum}
    \vspace{-0.2cm}
\end{equation}
in which $I(x_{InD}; z_{InD}|Y)$ and $I(z_{InD};Y)$ denote the \emph{Instance-Specific (IS)} and \emph{Class-Specific (CS) information} respectively. In \cref{fig:ana}, IS information is blue and orange areas, and CS information is yellow and green areas. CS information is for the closed-set label prediction task $Y$, while IS information is the special information of each sample that is not related to $Y$.

To analyze the information about OSAR, we let $T$ be a random variable that represents the task to distinguish OoD samples from InD samples, then we divide the information contained in $z_{InD}$ about $T$ into two parts~\cite{wang2022rethinking}:
\vspace{-0.2cm}
 \begin{equation}
I(z_{InD};T)=\underbrace{I(z_{InD}|Y; T)}_{IS \; \text{about} \; T} + \underbrace{I(z_{InD};Y;T)}_{CS \; \text{about} \; T},
     \label{eq:OSAR}
     \vspace{-0.2cm}
 \end{equation}
where $I(z_{InD}|Y; T)$ and $I(z_{InD};Y;T)$ are the information about the OoD detection task $T$ in IS and CS information (orange and green areas in \cref{fig:ana} respectively). We can see that larger IS and CS information are helpful for OSAR.

In this paper, we aim to enlarge the information about $T$ contained in CS and IS information for better OSAR performance, as illustrated in \cref{fig:1} (b) and the enlarged green and orange areas in \cref{fig:ana}. We first analyze the CS and IS information behaviors under the classical C.E. loss, and find that CS information is encouraged to be maximized but IS information tends to be eliminated in \cref{sec:CS_IS_CE}. Then we explain this conclusion from the IB theory view in \cref{sec:ib_analy}.

\subsection{CS and IS Information Behavior under C.E.}
\label{sec:CS_IS_CE}
CS information is for closed-set classification task $Y$, so it is similar for the same class sample, but distinct for the different class sample ($s_1,s_2/s_3$ in \cref{fig:1}). In contrast, IS information is not related to $Y$ and it is distinct for samples in the same class ($s_1, s_2$ in \cref{fig:1}). Therefore, we have the following proposition which describe the relation between CS/IS information and feature representation similarity.
\begin{proposition}
\vspace{-0.2cm}
For two feature representations of samples in the same class, more CS information means these two feature representations are more similar, and more IS information decreases their feature similarity.
\label{prop:cs_is}
\vspace{-0.2cm}
\end{proposition}

CS information is for the closed-set label prediction task $Y$, which is fully supervised by C.E. loss, so it is maximized during training.  In contrast, Eq.~\ref{eq:PL} shows that C.E. encourages representations of the same class to be exactly same with the corresponding prototype, and such high similarity eliminates the IS information according to Proposition~\ref{prop:cs_is}. Therefore, \textbf{C.E. loss tends to maximize the CS information and eliminate the IS information in the feature representation}. We analyze this conclusion based on Information Bottleneck (IB) theory in next~\cref{sec:ib_analy}.

\subsection{IB Theory Analysis for CS and IS Information}
\label{sec:ib_analy}

Applying the Data Processing Inequality~\cite{cover1999elements} to the Markov chain $Y \to x_{inD} \to z_{InD}$, we have
\vspace{-0.2cm}
\begin{equation}
    I(z_{InD};Y) \le I(x_{InD};Y).
    \label{eq:mar}
    \vspace{-0.2cm}
\end{equation}
It means that the compressed representation $z_{InD}$ cannot contain more information of $Y$ compared to the original data $x_{InD}$. 

According to the IB theory~\cite{tishby2000information,tishby2015deep}, the NN is to find the optimal solution of $z_{InD}$ with minimizing the following Lagrange:
\vspace{-0.2cm}
\begin{equation}
    \mathcal{L}[p(z_{InD}|x_{InD})]=I(z_{InD}; x_{InD})-\beta I(z_{InD}; Y),
    \label{eq:lar}
    \vspace{-0.2cm}
\end{equation}
where $\beta$ is the Lagrange multiplier attached to the constrained meaningful condition. Eq.~\ref{eq:lar} demonstrates the NN is solving a trade-off problem, as the first term tends to keep the information of $x_{InD}$ as less as possible while the second term tends to maximize the information of $Y$.

Inspired by \cite{wang2022rethinking,achille2018emergence}, the sufficient and minimum sufficient representation of $x_{InD}$ about $Y$ can be defined as:
\begin{definition}(Sufficient Representation) A feature representation $z_{InD}^{suf}$ of $x_{InD}$ is sufficient for $Y$ if and only if $I(z_{InD}^{suf};Y)=I(x_{InD}; Y)$.
\label{def:suf}
\end{definition}
\begin{definition}(Minimum Sufficient Representation) A sufficient representation $z_{InD}^{min}$ of $x_{InD}$ is minimum if and only if $I(z_{InD}^{min};x_{InD}) \le I(z_{InD}^{suf}; x_{InD})$, ${\forall} z_{InD}^{suf}$ that is sufficient for $Y$.
\label{def:mini}
\end{definition}

\noindent \textbf{CS Information Maximization.} The goal of training is to optimize $f$ so that $I(z_{InD};Y)$ (CS information) can approximate $I(x_{InD};Y)$, which stays unchanged as data distribution is fixed during training. Therefore, CS information is supposed to be maximized to the upper bound $I(x_{InD};Y)$ because of Eq.~\ref{eq:mar}. In this way, the closed-set classification task pushes the NN to learn the sufficient representation $z_{InD}^{suf}$ according to definition~\ref{def:suf}~\cite{federici2020learning}. 

\noindent \textbf{IS Information Elimination.}
When $z_{InD}$ is close to the sufficient representation $z_{InD}^{suf}$, the second term in Eq.~\ref{eq:lar} will be the fix value $I(x_{InD}; Y)$ based on the definition~\ref{def:suf}. So the key to minimize Eq.~\ref{eq:lar} is to minimize the first term $I(z_{InD}^{suf}; x_{InD})$. Based on the definition~\ref{def:mini}, the lower bound of $I(z_{InD}^{suf}; x_{InD})$ is $I(z_{InD}^{min}; x_{InD})$, so we can conclude that the learned representation is supposed to be the minimum sufficient representation $z_{InD}^{min}$~\cite{wang2022rethinking}. We substitute $I(z_{InD}^{suf}; x_{InD})$ and $I(z_{InD}^{min}; x_{InD})$ in definition~\ref{def:mini} with Eq.~\ref{eq:sum} and we have
\vspace{-0.2cm}
\begin{align}
    & I(x_{InD}; z_{InD}^{min}|Y) + I(z_{InD}^{min};Y) \nonumber \\
    \le &  
    I(x_{InD}; z_{InD}^{suf}|Y) + I(z_{InD}^{suf};Y).
    \label{eq:ine}
\end{align}
As both $z_{InD}^{min}$ and $z_{InD}^{suf}$ are sufficient, the second term of both sides in Eq.~\ref{eq:ine} is $I(x_{InD}; Y)$, so we have
\begin{equation}
    0 \le I(x_{InD}; z_{InD}^{min}|Y)
    \le
    I(x_{InD}; z_{InD}^{suf}|Y).
\end{equation}
Therefore, the learned IS information in $z_{InD}^{min}$ is smaller than any IS information in $z_{InD}^{suf}$, which could be eliminated to 0~\cite{wang2022rethinking} (no blue and orange areas in $z_{InD}^{min}$ in \cref{fig:ana}).

\subsection{Enlarge CS and IS Information for OSAR}
\label{sec:lar_cs_is}
Based on the analysis in \cref{sec:CS_IS_CE} and \cref{sec:ib_analy}, we show that C.E. tends to maximize the CS information and eliminate the IS information in the feature representation. Both larger IS and CS information are crucial for OSAR according to Eq.~\ref{eq:OSAR}, but C.E. does not bring the optimal information. On the one hand, IS information is eliminated so we lose a part of information which is beneficial for the OSAR. On the other hand, the learned representation is not sufficient and does not contain enough CS information in practice due to the model capacity and data distribution shift between training and test sets, which can be supported by the fact that test accuracy cannot reach 100\%. Therefore, we propose our method to enlarge the CS and IS information for better OSAR performance in next~\cref{sec:method}.

\section{Methods}
\label{sec:method}

\begin{figure*}[t]
    \centering
    \includegraphics[width=0.99\linewidth]{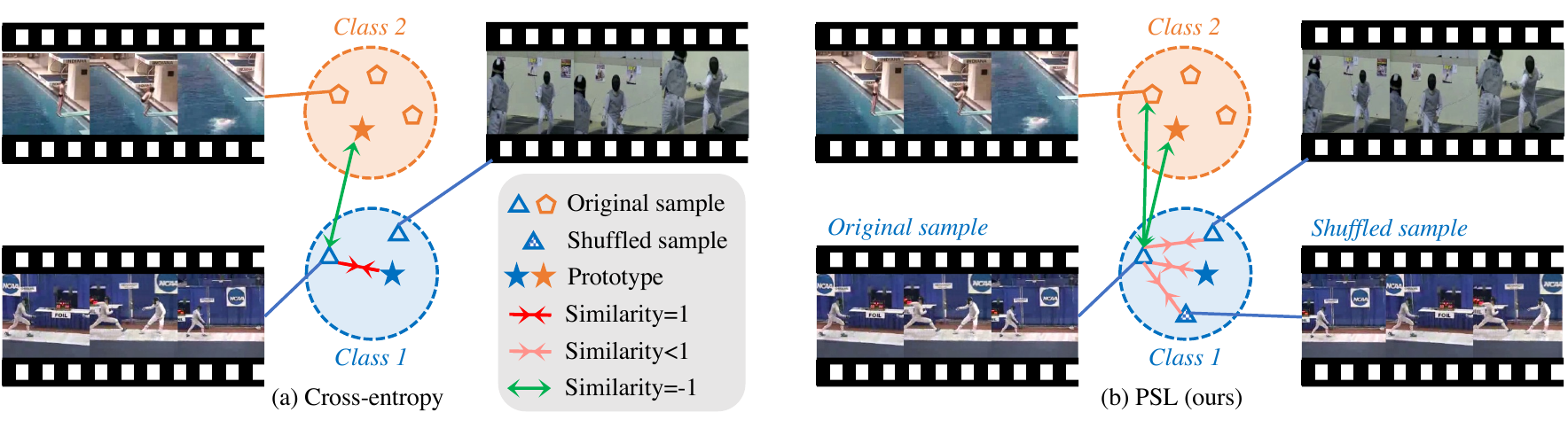}
    \vspace{-0.5cm}
    \caption{(a) C.E. encourages the sample feature $z$ to be exactly same with the corresponding prototype $k_i$. (b) Our PSL encourages the similarity between $z$ and $k_i$, features of shuffled sample $Q_{shuf}$ and other samples in the same class $Q_{sc}$ to have a similarity less than 1.}
    \label{fig:psl}
\end{figure*}

\subsection{Prototypical Similarity Learning}
\label{subsec:psl}

According to \cref{sec:CS_IS_CE}, we notice that IS information is suppressed by the C.E. loss and a key reason is C.E. encourages feature representations of the same class to be exactly same. Therefore, we argue that the feature representation of the same class samples should have a similarity $s < 1$. In other words, we aim to \textbf{keep the intra-class variance which prevents intra-class collapse to retain IS information}. Based on the classical PL loss Eq.~\ref{eq:PL}, we develop prototypical similarity learning (PSL):
\vspace{-0.2cm}
\begin{equation}
    \mathcal{L}_{PSL} = - \log \frac{\exp (\frac{1 - \left | z^T k_{i} - s \right |} {\tau}) }{ \exp (\frac{1 - \left | z^T k_{i} - s \right |} {\tau}) + \sum\limits_{n \in K_i^{-}} \exp (\frac{z^T  n} {\tau})},
    \label{eq:PSL}
    \vspace{-0.2cm}
\end{equation}
where $s$ and $\tau$ are fixed hyperparameters. In this way, we expect the prototype $k_i$ to act as the CS information for the InD class $i$, which is used to predict the label, and the dissimilarity between the $z$ and $k_i$ represents the IS information. Traditional PL loss (or C.E. loss) encourages the features of samples in the same classes to be as tight as possible, while our PSL aims to keep the variance within the same class. 

However, we find Eq.~\ref{eq:PSL} will converge to the trivial solution, where the $z$ converges to the training result of Eq.~\ref{eq:PL} and only $k_i$ shifts. To solve this problem, we introduce the similarity between different samples within a mini-batch into the denominator of Eq.~\ref{eq:PSL}. In this way, we directly constrain the relationship between sample features instead of only supervising the similarity between the sample feature and its prototype. We name the modified loss as PSL with contrastive terms (CT): 
\vspace{-0.2cm}
\begin{align}
\vspace{-0.2cm}
&\mathcal{L}_{PSL}^{CT} = \nonumber \\
&\frac{\exp (\frac{1 - \left | z^T k_{i} - s \right |} {\tau}) }{ \exp (\frac{1 - \left | z^T k_{i} - s \right |} {\tau}) + \sum\limits_{n \in Q_n} \exp (\frac{z^T  n} {\tau})+\sum\limits_{p \in Q_{sp} } \exp (\frac{\left | z^T p - s \right |} {\tau})},
    \label{eq:PSL_CT}
\end{align}
where $Q_n=K_i^{-} \cup Q_{ns}$. $Q_{ns}$ refers to the negative samples, \textit{i.e.}, samples in other classes, and $Q_{sp}$ refers to the soft positive samples which contains samples in the same class $Q_{sc}$ here. The reason we call soft positive samples is that we think samples in the same class share CS information but have distinct IS information.

\subsection{Video Shuffling for PSL}
\label{sec:shuf_PSL}

PSL aims to keep IS information during training, and in this section we introduce how to enlarge CS information through video shuffling. The appearance bias is a significant problem in the OSAR. For instance, the OoD classes \emph{Smile} and \emph{Chew} are easily classified as InD classes \emph{ApplyEyeMakeup} and \emph{ApplyLipstick}, as the majority area of all these classes are occupied by a face, as shown in Fig.~\ref{fig:ood_samples}. The NN is confused by the extremely similar spatial information and neglects the minor different temporal information. This phenomenon encourages us to strengthen the temporal information extraction ability of the NN to distinguish classes with very similar appearances but different actions. We find that introducing a simple yet effective way, \textit{i.e.}, to regard the shuffled video $Q_{shuf}$ as the soft positive sample in Eq.~\ref{eq:PSL_CT}, is extremely suitable and useful in our PSL framework. In this case, $Q_{sp}=Q_{sc} \cup Q_{shuf}$. Shuffled video means shuffling the frames within a single video. As the appearance information of the shuffled video is almost the same as the original video, a smaller than 1 similarity forces the NN to learn the distinct temporal information between them. Unlike existing works which predict the sequence or the type of the shuffled video~\cite{shi2022shuffle, jenni2020video, fernando2017self, lee2017unsupervised}, we regard the shuffle video as a whole sample and directly compare its feature representation with the original video in our PSL. \textbf{We find this technique can improve the closed-set accuracy which indicates more CS information is learned}. We summarize the difference between our PSL and classical C.E. in \cref{fig:psl}.

\subsection{Uncertainty Score}
\label{sec:uncer_score}
As our PSL aims to learn richer CS and IS information in the feature representation, we use the Mahalanobis distance to measure the uncertainty as it can be calculated from the feature representation perspective~\cite{sehwag2021ssd,lee2018simple}:
\begin{equation}
    u=(z-\mu_m)^{T}{\textstyle \sum_{m}^{-1}} (z-\mu_m),
    \label{eq:unce}
\end{equation}
where $\mu_m$ and ${\textstyle \sum_{m}}$ denote the mean and covariance of the whole training set features, and $z$ is the test sample feature.
\section{Experiments}
\label{sec:exp}

\noindent \textbf{Datasets.} Following~\cite{bao2021evidential}, we use UCF101~\cite{soomro2012ucf101} as the InD dataset for training and closed-set evaluation, and use HMDB51~\cite{kuehne2011hmdb} and MiT-v2~\cite{monfort2021multi} as OoD data for open-set evaluation. Different from~\cite{bao2021evidential} which does not clean the OoD data that may contains InD classes, we remove the overlapping classes between InD and OoD dataset during evaluation. See Appendix A for more details.

\noindent \textbf{Evaluation protocols.} For closed-set performance, we evaluate like the traditional way to calculate the top-1 accuracy Acc. (\%). For open-set performance, we follow the classical open-set recognition protocol~\cite{hendrycks2016baseline,hendrycks2018deep} to use the obtained uncertainty score Eq.~\ref{eq:unce} to calculate AUROC (\%), AUPR (\%) and FPR95(\%).\footnote{We find AUROC in~\cite{bao2021evidential} only considers one specific threshold based on their code, and after discussion and agreement they provide the modified correct score in our Tab.~\ref{tab:bench_hmdb}. See Appendix B for details.}

\begin{table*}[t]
\centering
\tablestyle{6pt}{0.6}
\begin{tabular}{llcccccccc}
\toprule[1pt]
& & \multicolumn{4}{c}{\bf{w/o K400 Pretrain}} & \multicolumn{4}{c}{\bf{w/ K400 Pretrain}} \\ \cmidrule{3-10}
   \bf{Datasets} &\bf{Methods}                     & \bf{AUROC$\uparrow$}    & \bf{AUPR$\uparrow$}  &\bf{FPR95$\downarrow$}  & \bf{Acc.$\uparrow$}  & \bf{AUROC$\uparrow$}   & \bf{AUPR$\uparrow$} &\bf{FPR95$\downarrow$}   &  \bf{Acc.$\uparrow$}  \\ \midrule
  \multirow{7}{*}{\makecell[l]{UCF101 (InD) \\ HMDB51 (OoD)}}
&OpenMax~\cite{bendale2016towards}     & \textit{82.28}    & \textit{54.59}   & \textit{50.69} & \textit{73.92} & 90.89  & 73.16   &38.77               & 95.32        \\
&MC Dropout~\cite{gal2016dropout}     & 75.75    & 41.21   & 54.78 & 73.63                              & 88.23   & 67.62   &38.12            & 95.06           \\
&BNN SVI~\cite{krishnan2018bar}      & 80.10    & 53.43  & 52.33  & 71.51                                & \textit{91.81}   & \textit{79.65}  &31.43             & 94.71           \\
&SoftMax~\cite{hendrycks2016baseline}    & 79.72    & 52.13  & 53.22  &\textit{73.92}                                   & 91.75   & 77.69  &\textit{28.60}              & 95.03           \\
&RPL~\cite{chen2020learning}        & 79.67    & 51.85  &56.40  & 71.46                                  & 90.53   & 77.86  &37.09               &\textit{95.59}           \\
&DEAR~\cite{bao2021evidential}      & 80.00         &49.23         &53.28 & 71.33                                    &84.16   & 75.54  &89.40                   & 94.48           \\
&PSL(ours)   & \bf{86.43}    & \bf{65.54}   & \bf{41.67} & \bf{76.53}                                  & \bf{94.05}   & \bf{86.55}   &\bf{23.18}   & \bf{95.62}           \\
&$\mathbf{\Delta}$ &\bf \textcolor{themeblue}{(+4.15)} &\bf \textcolor{themeblue}{(+10.95)} &\bf \textcolor{themeblue}{(-9.02)} &\bf \textcolor{themeblue}{(+2.61)} &\bf \textcolor{themeblue}{(+2.24)} &\bf \textcolor{themeblue}{(+6.90)} &\bf \textcolor{themeblue}{(-5.42)} &\bf \textcolor{themeblue}{(+0.03)}\\
\midrule
\multirow{7}{*}{\makecell[l]{UCF101 (InD) \\ MiTv2 (OoD)}} 
&OpenMax~\cite{bendale2016towards}       & \textit{84.43}   & \textit{76.69} &\textit{47.74}  & \textit{73.92}                 & \textit{93.34}   & 88.14   & \textit{28.95} & 95.32           \\
&MC Dropout~\cite{gal2016dropout}                                   & 75.66    & 62.20   & 51.57 & 73.63            & 88.71   & 83.36   & 39.46 & 95.06           \\
&BNN SVI~\cite{krishnan2018bar}                                      & 79.48    & 71.73   & 52.52 & 71.51            & 91.86   & \textit{90.12}   & 36.21 & 94.71           \\
&SoftMax~\cite{hendrycks2016baseline}                                      & 80.55    & 73.17   & 50.49 & \textit{73.92}            & 91.95   & 89.16   & 32.00 & 95.03           \\
&RPL~\cite{chen2020learning}                                          & 80.21    & 72.04   & 52.83 & 71.46            & 90.64   & 88.79   & 38.43 & \textit{95.59}           \\
&DEAR~\cite{bao2021evidential}                                         & 79.00          & 67.10        &52.44   &  71.33             & 86.04   & 87.38   & 87.40 & 94.48           \\
&PSL(ours)            & \bf{86.53}   & \bf{79.95}  & \bf{40.99} &  \bf{76.53}                & \bf{95.75}        & \bf{94.96}        & \bf{18.96} & \bf{95.62}                \\ 
&$\mathbf{\Delta}$ &\bf \textcolor{themeblue}{(+2.10)} &\bf \textcolor{themeblue}{(+3.26)} &\bf \textcolor{themeblue}{(-6.75)} &\bf \textcolor{themeblue}{(+2.61)} &\bf \textcolor{themeblue}{(+2.41)} &\bf \textcolor{themeblue}{(+4.84)} &\bf \textcolor{themeblue}{(-9.99)} &\bf \textcolor{themeblue}{(+0.03)} \\
\bottomrule[1pt]
\end{tabular}
\vspace{-0.3cm}
\caption{Comparison with state-of-the-art methods on {\bf HMDB51 and MiTv2 (OoD)} using TSM backbone. Acc. refers to closed-set accuracy. AUROC, AUPR and FPR95 are open-set metrics. Best results are in {\bf bold} and second best results in \textit{italic}. The gap between best and second best is in {\bf \textcolor{themeblue}{blue}}. DEAR and our methods contain video-specific operation.}
\label{tab:bench_hmdb}
\vspace{-0.6cm}
\end{table*}

\begin{figure}[t]
\centering
\includegraphics[width=0.48\textwidth]{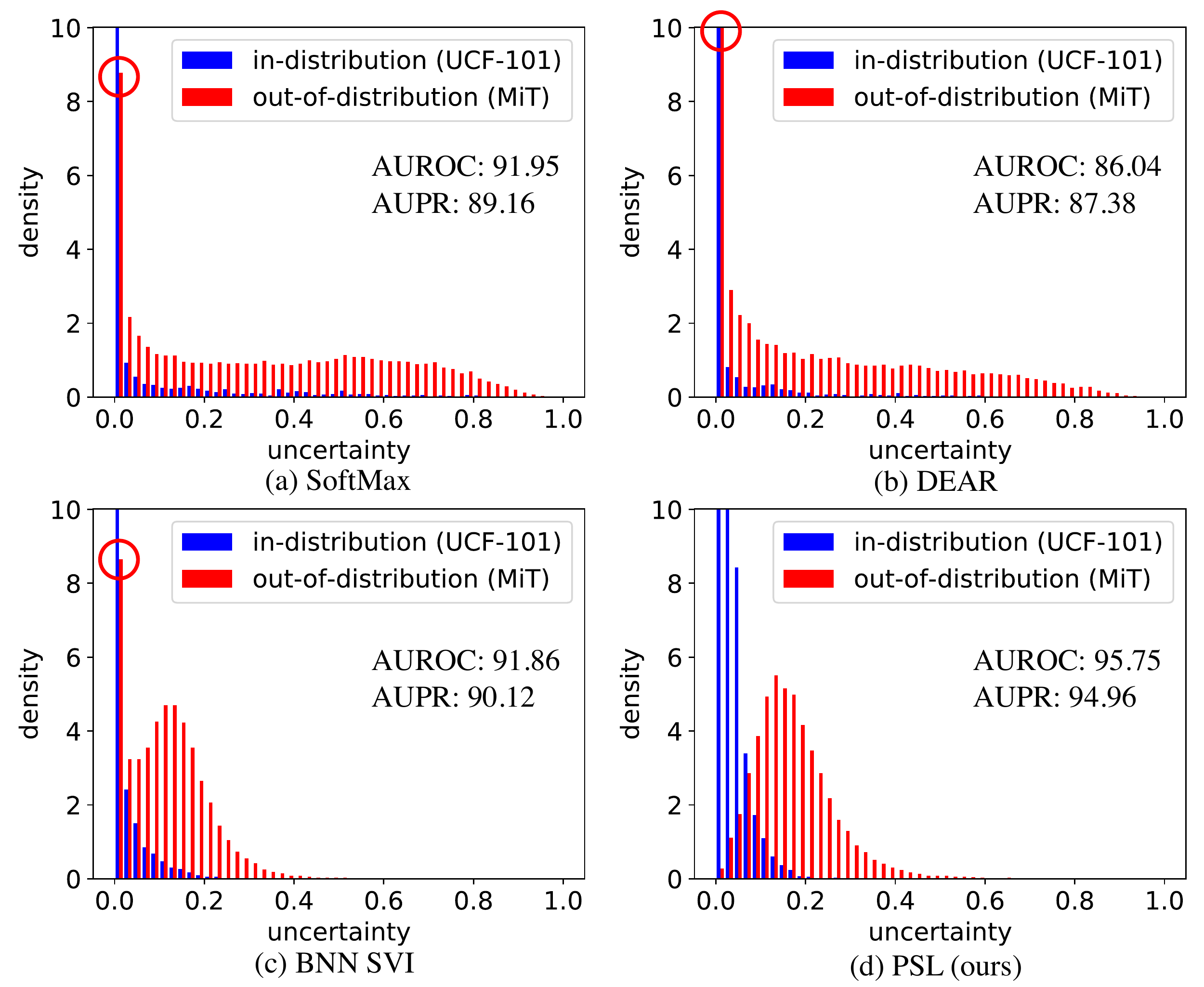}
\vspace{-8mm}
\caption{
The uncertainty distribution of InD and OoD samples of (a) Softmax, (b) DEAR, (c) BNN SVI and (d) our PSL method.
}
\label{fig:uncer_method}
\end{figure}

\noindent \textbf{Implementation details.} For Kinetics400 (K400)~\cite{i3d} pretrained model, our implementation setting is the same with~\cite{bao2021evidential}. The base learning rate is 0.001 and step-wisely decayed every 20 epochs with total of 50 epochs. We argue that as K400 is extremely large, the K400 pretrained model may already have seen the OoD data used in inference, so we conduct experiments from scratch (no ImageNet pretrained) to ensure that OoD data is absolutely unavailable during training. We use the LARS optimizer~\cite{you2017large} and set the base learning rate and momentum as 0.6 and 0.9 with total of 400 epochs. The experiments are conducted on TSM~\cite{lin2019tsm}, I3D~\cite{i3d} and SlowFast~\cite{feichtenhofer2019slowfast}. The batch size for all methods is 256. More details are in Appendix C.

\subsection{Evaluation Results}
\label{sec:res}

\begin{figure}[t]
\centering
\includegraphics[width=0.48\textwidth]{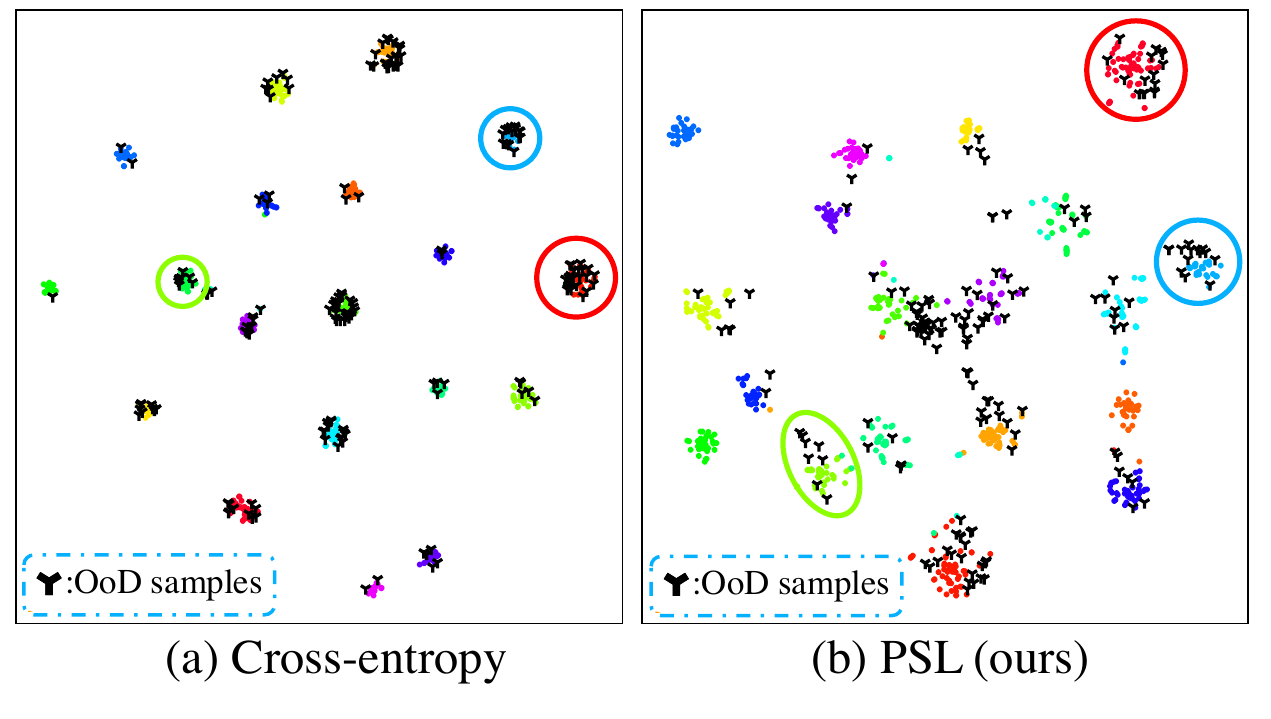}
\vspace{-10mm}
\caption{
Feature representation visualization of cross-entropy and our PSL method. OoD samples are in black and InD samples are in other colors. In the red, blue and green circles, it is clear that OoD samples distribute at the edge of InD samples in our PSL, while greatly overlap with each other in the cross-entropy method.
}
\label{fig:tsne}
\end{figure}

\noindent \textbf{Comparison with state-of-the-art.} We report the results on HMDB51 (OoD) and MiT-v2 (OoD) in Table~\ref{tab:bench_hmdb} using TSM backbone~\cite{lin2019tsm}. The evaluation results of other backbones including I3D and SlowFast are in the Appendix D. We can see that for w/ or w/o K400 pretrain, our PSL method has significantly better open-set and closed-set performance than all baselines. The uncertainty distribution of InD and OoD samples are depicted in Fig.~\ref{fig:uncer_method} for MiT-v2 (OoD) with K400 pretrained. Three baseline methods have a clear over confidence problem, \textit{i.e.}, the far left column is extremely high (red circles in Fig.~\ref{fig:uncer_method}), which means a large number of OoD samples have almost 0 uncertainty, while our method significantly alleviates this problem through the distinct representation of OoD samples, illustrated in Fig.~\ref{fig:tsne}. Besides, we can find that the open-set performance w/ K400 pretrain is higher than w/o pretrain for almost all methods in Table~\ref{tab:bench_hmdb} and \cref{fig:1} (a), which can testify the importance of richer semantic representation for OSAR.

\begin{table*}[t!]
\tablestyle{6pt}{0.3}
\centering
\begin{tabular}{lcccccccccccc}
\toprule[1pt]
&&&&& \multicolumn{2}{c}{\bf{InD}} &\multicolumn{2}{c}{\bf{OoD}} \\ \cmidrule{6-9}
                        & \bf{$s$} & \bf{$Q_{ns}$} & \bf{$Q_{sc}$} & $Q_{shuf}$ &\bf{Mean} & \bf{Variance} &\bf{Mean} & \bf{Variance} & \bf{AUROC$\uparrow$} & \bf{AUPR$\uparrow$} &\bf{FPR95$\downarrow$}  & \bf{Acc.$\uparrow$} \\ \midrule
$\mathcal{L}_{PL}$                   & \XSolidBrush  & \XSolidBrush      & \XSolidBrush      & \XSolidBrush        &0.81 &0.0015 &0.63 &0.0029 & 80.95 & 52.79 & 52.51 & 72.36          \\ \midrule
$\mathcal{L}_{PSL}$ & \Checkmark  & \XSolidBrush      &\XSolidBrush       & \XSolidBrush        & 0.79     &0.0016  & 0.62 & 0.0028 & 81.79 & 54.16 & 52.33 & 72.33          \\ \midrule
 \multirow{3}{*}{$\mathcal{L}_{PSL}^{CT}$}                       & \Checkmark  & \Checkmark      & \XSolidBrush      & \XSolidBrush        &0.71 &0.0022 &0.61 &0.0036 & 82.60 & 57.36 & 50.03 & 72.17          \\ 
                        & \Checkmark  & \Checkmark      & \Checkmark      & \XSolidBrush        & 0.71 &0.0023 &0.49 &0.0035 & 83.42 & 59.05 & 51.32 & 72.28          \\
                        & \Checkmark  & \Checkmark      & \Checkmark      & \Checkmark        & 0.74 &0.0016 &0.63 &0.0029 & 86.43 & 65.58 & 41.75 & 77.19     \\ 
                        \bottomrule[1pt]  
\end{tabular}
\vspace{-0.3cm}
\caption{Abaltion results of different components in $\mathcal{L}_{PSL}^{CT}$.}
\label{tab:abla}
\vspace{-0.5cm}
\end{table*}

\noindent \textbf{Comparison with metric learning methods.} Our method concentrates on the feature representation aspect for the OSAR problem, so we also implement several well-known metric learning methods and show the result in Table~\ref{tab:metric_learning}. The evaluation is conducted using TSM model and OoD dataset is HMDB51. We do not use video shuffling in our method for fair comparison. We can see that our method still achieves the best open-set performance. The most important difference between our method and all other metric learning methods is that they aim to push the features of one class as tight as possible like C.E., while our method aims to keep the feature variance within a class to retain IS information. We calculate the mean similarity between the sample feature and the corresponding class center. The mean similarity ranges from 0.77 to 0.82 for other metric learning methods, while mean similarity is 0.71 ($s=0.8$) and 0.6 ($s=0.6$) for our PSL. So our method has looser feature distribution within a class, as shown in \cref{fig:tsne}.

\begin{table}[t!]
\tablestyle{5pt}{0.8}
\centering
\begin{tabular}{lcccccccccc}
\toprule[1pt]
& \bf{AUROC$\uparrow$} & \bf{AUPR$\uparrow$} &\bf{FPR95$\downarrow$}  & \bf{Acc.$\uparrow$} \\ \midrule
SoftMax    &80.95	&52.79	&52.51   &72.36	        \\
 Triplet \cite{triplet} &81.02 &54.75	&53.88 &75.50\\
Normface \cite{normface} &80.99	&54.90	&53.19 &73.34            \\
Circle  \cite{circle} 	&78.76	&51.65	&55.27 &72.15 	\\
Arcface \cite{arcface} &81.23	&55.03	&53.67 &\bf{75.95}	\\
LSoftMax \cite{lsoftmax} &80.87 &54.01	&52.29 &73.05 \\ \midrule
PSL($s=0.8$)   & \bf{83.42} & \bf{59.05} & \bf{51.32} & 72.28           \\
PSL($s=0.6$)    &82.75 & 58.57 & 52.27 & 73.26           \\
                        \bottomrule[1pt]  
\end{tabular}
\vspace{-0.3cm}
\caption{Comparison with different metric learning methods.}
\label{tab:metric_learning}
\vspace{-0.4cm}
\end{table}

\subsection{Ablation Study}
\label{sec:abl_stu}

\begin{figure}[t]
    \centering
    \includegraphics[width=0.99\linewidth]{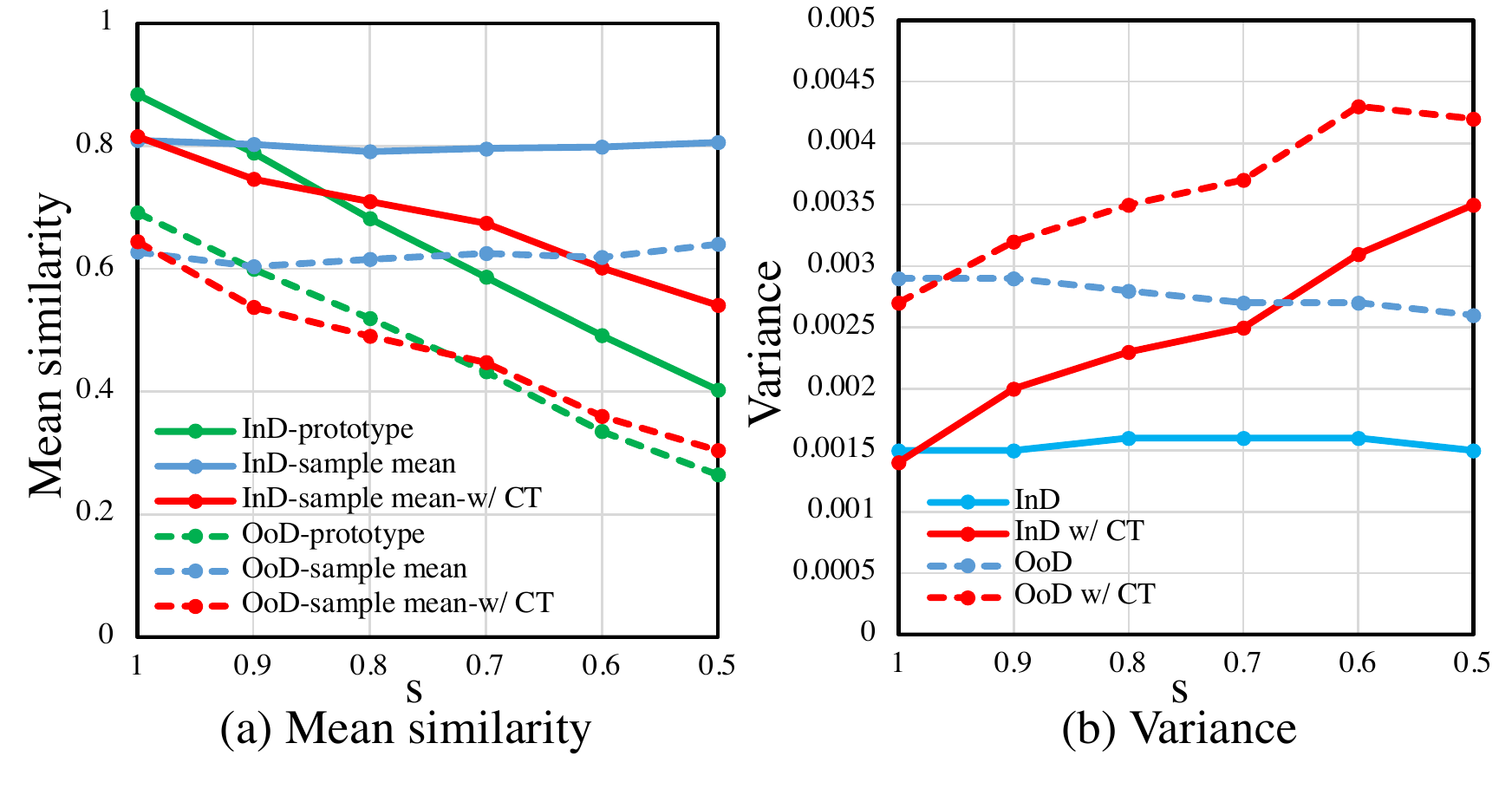}
    \vspace{-0.5cm}
    \caption{Mean similarity and variance analysis for CT terms.}
    \label{fig:abl_contra}
    \vspace{-0.3cm}
\end{figure}

\noindent \textbf{Contrastive terms in $\mathcal{L}_{PSL}^{CT}$ for IS information.} The intuition of PSL is to keep the intra-class variance to retain the IS information which is helpful for OSAR. We expect that the representation $z$ within a class has a similarity $s<1$ with the prototype $k_i$, so each sample can keep its own IS information. However, we find that the loss $\mathcal{L}_{PSL}$ may lead the network to find the trivial representation of samples $z$ which is similar to using loss $\mathcal{L}_{PL}$, where only $k_i$ shifts and $z$ does not. We calculate the mean of similarity $sim(z, \bar z_i)$, where $\bar z_i$ denotes the mean representation of all samples in the same class $i$, and the mean of similarity with the corresponding prototype $sim(z, k_i)$, as well as the feature variance in all dimensions. Fig.~\ref{fig:abl_contra} (a) and (b) show that with the hyper-parameter $s$ decreasing, the $sim(z, k_i)$ decreases as expected by $\mathcal{L}_{PSL}$ (green curves), but the $sim(z, \bar z_i)$ and variance stay unchanged (blue curves), meaning that the representation of samples are still similar with using $\mathcal{L}_{PL}$, and only the prototypes are pushed away by the sample representations. In contrast, with CT in $\mathcal{L}_{PSL}^{CT}$, the $sim(z, \bar z_i)$ decreases and variance increases with $s$ decreases (red curves), indicating that CT is significantly effective to keep the intra-class variance.

To individually study the effectiveness of $Q_{ns}$ and $Q_{sc}$ in $\mathcal{L}_{PSL}^{CT}$, we provide the ablation results in Table~\ref{tab:abla}. For OoD samples, we calculate the similarity with the mean representation of its predicted class. Table~\ref{tab:abla} shows that using $Q_{ns}$ alone can significantly increase the intra-class variance for both InD and OoD samples, meaning the pushing effect of representations in other classes can implicitly help retain the IS information. On top of that, $Q_{sc}$ can further learn more IS information that is helpful to distinguish OoD samples, as the mean similarity of InD samples stay unchanged, but OoD samples are smaller which means OoD samples are far away from InD samples.

\noindent \textbf{Shuffled videos for CS information.} \cref{tab:abla} shows that $Q_{shuf}$ can improve both closed-set and open-set performance, which proves introducing shuffled videos in PSL can enlarge CS information. Smaller intra-class variance brought by $Q_{shuf}$ testify Proposition~\ref{prop:cs_is} that more CS information means more similar features within the same class.

\begin{figure}[t]
    \centering
    \includegraphics[width=0.99\linewidth]{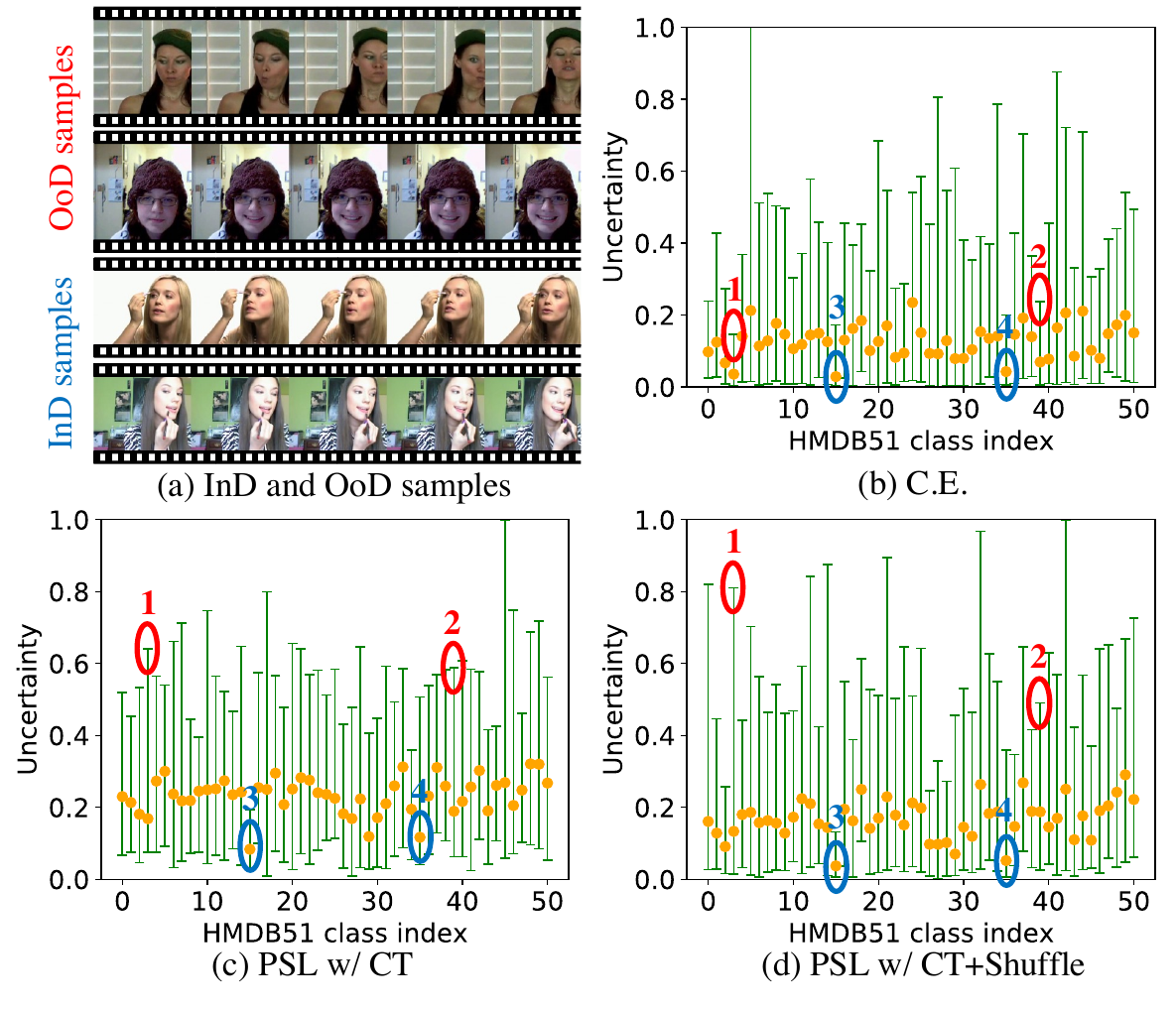}
    \vspace{-0.5cm}
    \caption{(a) \emph{chew} and \emph{smile} are OoD samples from HMDB51, and \emph{ApplyEyeMakeup} and \emph{ApplyLipstick} are InD samples from UCF101. (b-d) Uncertainty distribution of each class in HMDB51. Class 1: \emph{chew}, 2: \emph{smile}, 3: \emph{golf}, 4: \emph{shoot bow}. Classes 1 and 2 are OoD while 3 and 4 are InD.}
    \label{fig:ood_samples}
    \vspace{-0.4cm}
\end{figure}

We draw the uncertainty of all classes in HMDB51, as shown in Fig.~\ref{fig:ood_samples}. Note that some classes in HMDB51 are actually InD as they appear in the UCF101, like the class 3 \emph{golf} and 4 \emph{shoot bow} in Fig.~\ref{fig:ood_samples}. We find that in C.E. some OoD classes have extremely low uncertainty, such as class 1 \emph{chew} and 2 \emph{smile}, because they are spatially similar to some InD classes like \emph{ApplyEyeMakeup} and \emph{ApplyLipstick} in Fig.~\ref{fig:ood_samples} (a). Comparing (b) and (c) shows that our PSL can increase the average uncertainty of OoD classes (higher yellow points), and some OoD classes which are similar to InD classes like 1 and 2 have much higher uncertainty in our PSL method. After shuffled samples are involved, some InD classes whose uncertainty are increased in (c) like 3 and 4 have lower uncertainty in (d), and the uncertainty of some OoD classes sharing similar appearance with InD classes like class 1 is further improved.

$Q_{sp}$ in Eq.~\ref{eq:PSL_CT} contains $Q_{shuf}$ and $Q_{sc}$, so we analyze whether should we assign the same $s$ for the shuffled video $Q_{shuf}$ and other videos in the same class $Q_{sc}$. \cref{tab:s_q_shu} shows that the same $s$ have good enough performance. So we set the same $s$ for $Q_{shuf}$ and $Q_{sc}$ in the default setting to reduce the number of hyper-parameters.

\subsection{Discussion}
\label{sec:discussion}

\begin{table}[t]
\tablestyle{4pt}{0.8}
\centering
\begin{tabular}{ccccccc}
\toprule[1pt]
 $s(Q_{shuf})$ &$s(Q_{sc})$  &\bf{AUROC$\uparrow$} & \bf{AUPR$\uparrow$} &\bf{FPR95$\downarrow$}  & \bf{Acc.$\uparrow$} \\ \midrule
0.7 &\multirow{4}{*}{0.7} & 85.25	&63.91	&48.34 &76.98\\
0.5	& &86.03	&64.36	&43.70 & 76.53\\
0.3	& &83.80	&60.42	&48.76 & 75.50\\
0 & &79.54&50.59&54.43& 72.59\\ \midrule
0.8 &0.8 &86.43	&65.58	&41.75 &76.53\\
0.9 &0.9 &83.12	&57.04	&46.84 &73.31\\
1 &1 &82.04	&53.82&51.82 &72.89	\\
\bottomrule[1pt]
\end{tabular}
\vspace{-0.3cm}
\caption{Ablation study of similarity $s$ for $Q_{shuf}$ and $Q_{sc}$.}
\label{tab:s_q_shu}
\vspace{-0.3cm}
\end{table}

\begin{figure}[t]
    \centering
    \includegraphics[width=0.99\linewidth]{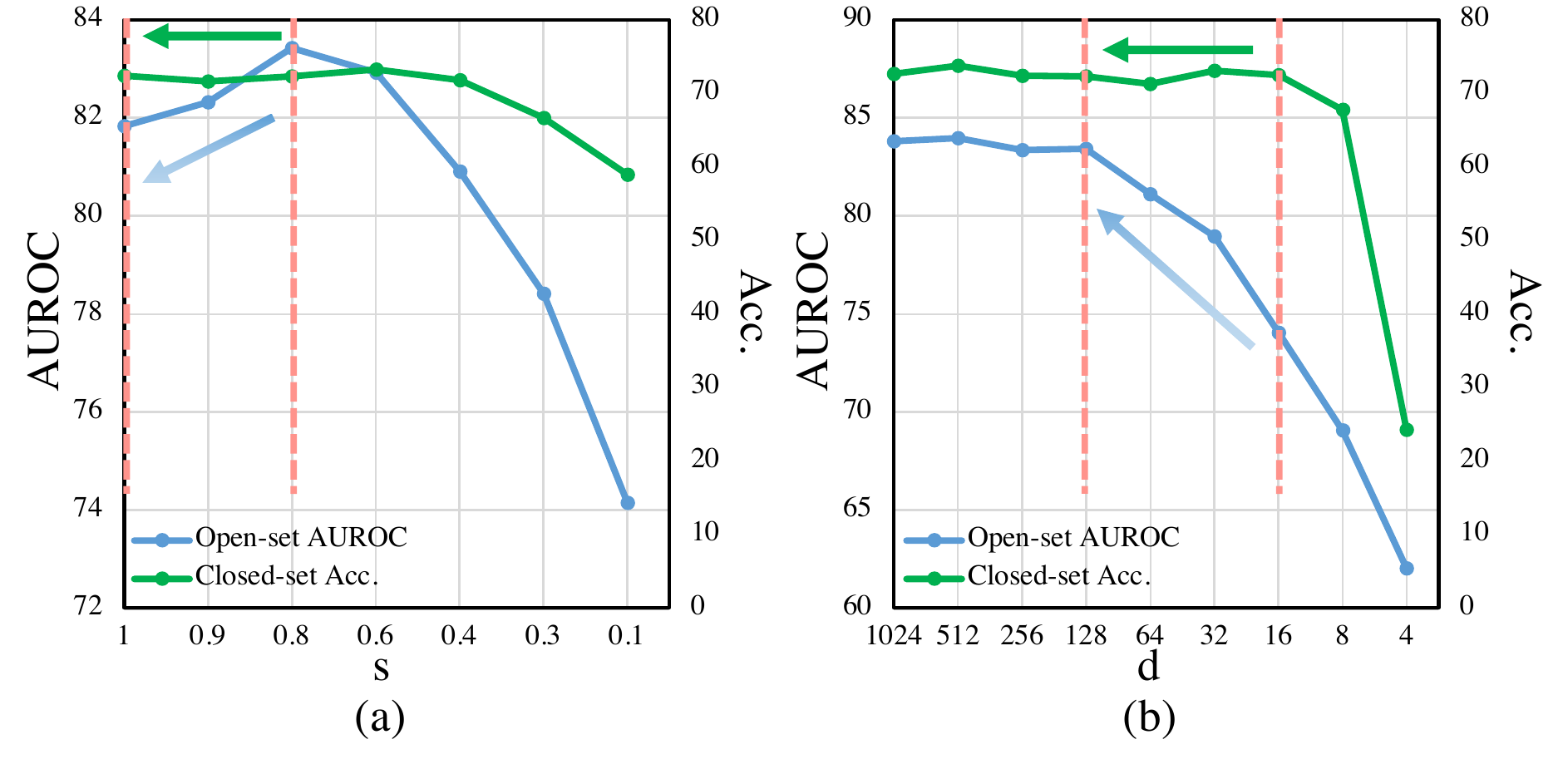}
    \vspace{-0.5cm}
    \caption{Ablation study of similarity $s$ and feature dimension $d$.}
    \label{fig:dis_s}
    \vspace{-0.4cm}
\end{figure}

\noindent \textbf{Both CS and IS information are useful.} We provide the closed-set and open-set performance under different hyper-parameter $s$ and feature dimension $d$ in \cref{fig:dis_s}. (a) shows that $s=0.8$ has better open-set performance than $s=1$ and has comparable closed-set accuracy, which illustrates that retaining the IS information which is eliminated by C.E. ($s=1$) is beneficial. When $s<0.8$, the NN cannot learn enough CS information, so both closed-set and open-set performance drops. Therefore, a proper mixture of CS and IS information is ideal. (b) shows that when $d$ grows from 4 to 16, more CS information is contained so that both closed-set and open-set performance improves. When $d$ grows from 16 to 128, the feature does not include more CS information as closed-set accuracy is comparable. However, open-set performance keeps increasing which means more IS information is contained based on more feature dimensions. This interesting experiment shows that enough information for closed-set recognition is not enough for open-set recognition because IS information is not related to the closed-set task but useful for the open-set task.

\begin{table}[t]
\tablestyle{2pt}{0.8}
\centering
\begin{tabular}{lcccccccccccc}
\toprule[1pt]
   \bf{Epoch}                      &\bf{Mean} & \bf{Variance} & \bf{AUROC$\uparrow$} & \bf{Acc-Test.$\uparrow$} & \bf{Acc-Train.$\uparrow$}\\ \midrule
200 &0.577 &3.3e-3 &75.08	&68.39 &99.85\\
400 &0.602 &3.1e-3 &82.92	&73.26 &100\\
800 &0.613 &3.0e-3 &82.54	&73.29 &100\\
                        \bottomrule[1pt]  
\end{tabular}
\vspace{-0.3cm}
\caption{Training process analysis when $s=0.6$ w/o $Q_{shuf}$.}
\label{tab:continual_train}
\vspace{-0.3cm}
\end{table}

\noindent \textbf{Feature variance and open-set performance analysis.} \cref{fig:dis_s} (a) shows that when features get looser ($s=1-0.8$), the open-set performance is improved, but if features get continually looser ($s=0.8-0.1$), the open-set performance drops. So there is no strict relation between the feature variance and open-set performance. One may argue that continual training can benefit the open-set performance~\cite{vaze2021open}, which is alongside with smaller feature variance~\cite{han2021neural}. We show that the benefit of continual training comes from better closed-set performance, not tighter features. \cref{tab:continual_train} shows that when we train the model from 200 to 400 epochs, the closed-set accuracy is higher, and feature is tighter (larger mean similarity and smaller variance), and the open-set performance is better. But from epoch 400 to 800 we find the model is already overfitted to the training set, as the accuracy of test set remains unchanged. So although the features get tighter in the 800 epoch, both the closed-set and open-set performance remain same.

\section{Conclusion}
We analyze the OSAR problem from the information perspective, and show that cross-entropy tends to eliminate IS information and cannot fully learns CS information which are both useful for the open-set task. So we propose PSL to retain IS information and introduce shuffle videos into PSL to enlarge CS information. Comprehensive experiments demonstrate the effectiveness of our PSL and the importance of IS and CS information in the OSAR task.

\noindent {\bf Acknowledgements} This work is supported by Alibaba Group through Alibaba Research Intern Program.

\appendix
\onecolumn
\begin{center}
\Large
\textbf{Appendices}
\end{center}
\section{Datasets}
We follow the datasets setting in~\cite{bao2021evidential}. The training InD dataset is UCF101, which contains 101 classes with 9537 training samples and 3783 test samples. The OoD datasets for open-set evaluation are HMDB51 and MiT-v2. We use the test sets of them which contain 1530 samples and 30500 samples respectively. For UCF101 and HMDB51, we follow the MMAction~\cite{mmaction2019} to use the split 1 for training and evaluation, which is the same with~\cite{bao2021evidential}. Note that in~\cite{bao2021evidential}, they find some classes in HMDB51 overlap with those in UCF101 but they do not clean them. We remove the overlapping classes in UCF101 and HMDB51 so that OoD data does not contain any samples of InD classes. The classes we remove in HMDB51 and the corresponding same classes in UCF101 are in Table~\ref{tab:dataset}.

\begin{table*}[h!]
\tablestyle{3pt}{1}
\centering
\begin{tabular}{lcccc}
\toprule[1pt]
HMDB51 &35, Shoot bow &29, Push up &15, Golf  &26, Pull up \\
UCF101 &2, Archery &71, PushUps &32, GolfSwing &69, PullUps \\ \midrule
HMDB51 &30, Ride bike &34, Shoot ball &43, Swing baseball &31, Ride horse \\
UCF101 &10, Biking &7, Basketball &6, BaseballPitch &41, HorseRiding \\
                        \bottomrule[1pt]  
\end{tabular}
\vspace{-0.3cm}
\caption{Overlapping classes in HMDB51 and UCF101.}
\label{tab:dataset}
\end{table*}
\section{Evaluation protocols}

Based on codes provided by~\cite{bao2021evidential}, we find that their evaluation metrics including Open maF1 and AUORC are both calculated under a specific certain threshold, \textit{i.e.}, a sample whose uncertainty is larger than the threshold will be considered as an OoD sample. The threshold is determined by top 5\% uncertainty in the training set. This is contradictory with the classical metrics in the open-set image recognition, in which common metrics including AUROC and AUPR~\cite{hendrycks2016baseline,hendrycks2018deep} both consider all thresholds. Each point on the ROC and PR curve is based on one specific threshold, and the area under ROC and PR curve is regarded as the comprehensive result of all thresholds. After discussing with authors in~\cite{bao2021evidential}, they admit that the AUROC, AUPR and FPR95 which are served as the classical metrics in the open-set image recognition are more suitable for the OSAR problem. So they modify the corresponding code and we provide the correct results in the Table 1 in our paper. We provide a comparison between the result of considering only one threshold and all thresholds in Table~\ref{tab:bench_hmdb_o}. The results show that no matter for only considering one threshold or all thresholds, our PSL method can both outperform all methods.
\begin{table*}[h!]
\centering
\tablestyle{3pt}{0.8}
\begin{tabular}{llcccccccc}
\toprule[1pt]
& & \multicolumn{4}{c}{\bf{One threshold}~\cite{bao2021evidential}} & \multicolumn{4}{c}{\bf{All thresholds (ours)}} \\ \cmidrule{3-10}
   \bf{Models} &\bf{Methods}                     & \bf{AUROC$\uparrow$}    & \bf{AUPR$\uparrow$}  &\bf{FPR95$\downarrow$}  & \bf{Acc.$\uparrow$}  & \bf{AUROC$\uparrow$}   & \bf{AUPR$\uparrow$} &\bf{FPR95$\downarrow$}   &  \bf{Acc.$\uparrow$}  \\ \midrule
  \multirow{7}{*}{TSM}
&OpenMax     & \underline{84.18}  &  \underline{76.52}  &  100   & 95.32      &90.89  & 73.16 &38.77   &95.32  \\
&MC Dropout     & 78.50  &  71.11  &   37.80 &95.06                          & 88.23   & 67.62   &38.12            & 95.06           \\
&BNN SVI      &  77.77  &  71.00  &   41.13   &       94.71                              & \underline{91.81}   & \underline{79.65}  &31.43             & 94.71           \\
&SoftMax    & 82.77  &  74.33  &  \underline{29.58}   &       95.03                                    & 91.75   & 77.69  &\underline{28.60}              & 95.03           \\
&RPL        & 77.75  &  70.93  &  40.87   &       \underline{95.59}                                 & 90.53   & 77.86  &37.09               &\underline{95.59}           \\
&DEAR      & 82.73  &  74.79  &  100   &       94.48                                  &84.16   & 75.54  &89.40                   & 94.48           \\
&PSL(ours)   & \bf{87.53}  &  \bf{79.92}  &   \bf{14.98}   &       \bf{95.62}                                  & \bf{94.05}   & \bf{86.55}   &\bf{23.18}   & \bf{95.62}           \\
&$\mathbf{\Delta}$ &\bf \textcolor{themeblue}{(+3.35)} &\bf \textcolor{themeblue}{(+3.10)} &\bf \textcolor{themeblue}{(-14.60)} &\bf \textcolor{themeblue}{(+0.03)} &\bf \textcolor{themeblue}{(+2.24)} &\bf \textcolor{themeblue}{(+6.90)} &\bf \textcolor{themeblue}{(-5.42)} &\bf \textcolor{themeblue}{(+0.03)}\\
\bottomrule[1pt]
\end{tabular}
\vspace{-0.3cm}
\caption{Comparison of different evaluation metrics on HMDB51 (OoD) with K400 pretrained.}
\label{tab:bench_hmdb_o}
\end{table*}

When we use MiT-v2 as the OoD dataset, we find the imbalance problem, which is also mentioned in~\cite{bao2021evidential}. The MiT-v2 test set contains 30500 samples while UCF101 test set only contains 3783 samples. This will cause the AUPR to be close to 100\% if we regard all samples in MiT-v2 as OoD samples during evaluation. Therefore, we divide the MiT-v2 test set into 10 splits, and evaluate the open-set metrics for 10 times and calculate the mean as the final result. A comparison between the results of evaluating 10 times and 1 time is shown in Table~\ref{tab:bench_mit_10}. The results illustrate that when we use all samples in MiT-v2 for open-set evaluation, the AUPR will be close to 100\%, although our method still achieves the best performance. The AUROC and FPR95 are not sensitive to the OoD sample numbers.

\begin{table*}[h!]
\centering
\tablestyle{3pt}{0.8}
\begin{tabular}{llcccccccc}
\toprule[1pt]
& & \multicolumn{4}{c}{\bf{1 time}} & \multicolumn{4}{c}{\bf{10 times}} \\ \cmidrule{3-10}
   \bf{Models} &\bf{Methods}                     & \bf{AUROC$\uparrow$}    & \bf{AUPR$\uparrow$}  &\bf{FPR95$\downarrow$}  & \bf{Acc.$\uparrow$}  & \bf{AUROC$\uparrow$}   & \bf{AUPR$\uparrow$} &\bf{FPR95$\downarrow$}   &  \bf{Acc.$\uparrow$}  \\ \midrule
  \multirow{7}{*}{TSM}
&OpenMax       & \underline{93.34}  &  98.46  &   \underline{29.20}   & 95.32                 & \underline{93.34}  &  88.14  &   \underline{28.95}   & 95.32           \\
&MC Dropout                                   & 88.71   & 97.92  & 39.46 & 95.06            & 88.71   & 83.36   & 39.46 & 95.06           \\
&BNN SVI                                      & 91.86   & \underline{98.75}   & 36.21 & 94.71            & 91.86   & \underline{90.12}   & 36.21 & 94.71           \\
&SoftMax                                      & 91.95   & 98.68   & 32.00 & 95.03            & 91.95   & 89.16   & 32.00 & 95.03           \\
&RPL                                          & 90.64   & 98.57   & 38.43 & \underline{95.59}            & 90.64   & 88.79   & 38.43 & \underline{95.59}           \\
&DEAR                                         & 86.04   & 98.08   & 87.66 & 94.48             & 86.04   & 87.38   & 87.40 & 94.48           \\
&PSL(ours)            & \bf{95.75}        & \bf{99.39}        & \bf{19.00} & \bf{95.90}                & \bf{95.75}        & \bf{94.96}        & \bf{18.96} & \bf{95.90}                \\ 
&$\mathbf{\Delta}$ &\bf \textcolor{themeblue}{(+2.41)} &\bf \textcolor{themeblue}{(+0.64)} &\bf \textcolor{themeblue}{(-10.20)} &\bf \textcolor{themeblue}{(+0.31)} &\bf \textcolor{themeblue}{(+2.41)} &\bf \textcolor{themeblue}{(+4.84)} &\bf \textcolor{themeblue}{(-9.99)} &\bf \textcolor{themeblue}{(+0.31)} \\
\bottomrule[1pt]
\end{tabular}
\vspace{-0.3cm}
\caption{Comparison of different evaluation methods on MiT-v2 (OoD) with K400 pretrained.}
\label{tab:bench_mit_10}
\end{table*}

\section{Implementation details}

When we use K400 pretrained model, the only method we need to fulfill is our PSL method, and we follow~\cite{bao2021evidential} to set the base learning rate as 0.001 and step-wisely decayed every 20 epochs with total 50 epochs. When we train the model from scratch, we need to conduct experiments on all methods in our Table 1. For our PSL method, we use the LARS optimizer~\cite{you2017large} and set the base learning rate and momentum as 0.6 and 0.9 with totally 400 epochs. The reason we use this strategy is inspired by the contrastive learning SimCLR~\cite{chen2020simple}. For other baselines, we find the above learning rate strategy cannot achieve good enough closed-set performance, and we find that setting the base learning rate as 0.05 and step-wisely decayed every 160 epochs with totally 400 epochs can achieve comparable closed-set performance. The batch size for all methods is 256, and we use 16 NVIDIA V100 GPUs to train the model.

\section{OSAR performance under I3D and SlowFast backbone}

We provide the OSAR results under TSM~\cite{lin2019tsm} backbone in Table 1 of the paper. Here, we further provide the OSAR results under I3D~\cite{i3d} and SlowFast~\cite{feichtenhofer2019slowfast} backbones in Table~\ref{tab:bench_i3d} and~\ref{tab:bench_slf}. We can see our PSL method still achieves state-of-the-art performance under these two backbones. The performance gain under Slowfast when MiTv2 is OoD dataset is marginal, as baselines already have high performance.

\begin{table*}[h!]
\centering
\tablestyle{3pt}{0.8}
\begin{tabular}{llcccccccc}
\toprule[1pt]
& & \multicolumn{4}{c}{\bf{w/o K400 Pretrain}} & \multicolumn{4}{c}{\bf{w/ K400 Pretrain}} \\ \cmidrule{3-10}
   \bf{Datasets} &\bf{Methods}                     & \bf{AUROC$\uparrow$}    & \bf{AUPR$\uparrow$}  &\bf{FPR95$\downarrow$}  & \bf{Acc.$\uparrow$}  & \bf{AUROC$\uparrow$}   & \bf{AUPR$\uparrow$} &\bf{FPR95$\downarrow$}   &  \bf{Acc.$\uparrow$}  \\ \midrule
  \multirow{7}{*}{\makecell[l]{UCF101 \\ HMDB51 }}
&OpenMax  &\underline{83.78}&\underline{54.65}&\underline{47.60}&\underline{74.42} & 92.03 & 77.72 &41.02 & \underline{95.01}\\ 
&MC Dropout &75.85&40.04&50.34&74.39 &91.66 &78.87 &33.60 &94.11\\
&BNN SVI &81.53&53.62&49.18&73.15 &91.57 &78.65 &34.60 &93.89\\
&SoftMax &81.24&54.21&48.20&\underline{74.42} &91.28 &79.73 &34.18 &94.11\\
&RPL &79.80&52.09&54.07&71.62 &\underline{92.49} &\underline{81.72} &\underline{28.89} &94.26\\
&DEAR &78.91&54.14&81.96&\underline{74.42} &89.80 &80.86 &75.63 &93.89\\
&PSL(ours) &\bf{86.88}&\bf{65.63}&\bf{39.85} &\bf{78.85} &\bf{93.62} &\bf{85.54} &\bf{28.38} &\bf{95.46}\\
&$\mathbf{\Delta}$ &\bf \textcolor{themeblue}{(+3.10)} &\bf \textcolor{themeblue}{(+10.98)} &\bf \textcolor{themeblue}{(-7.75)} &\bf \textcolor{themeblue}{(+4.43)} &\bf \textcolor{themeblue}{(+1.13)} &\bf \textcolor{themeblue}{(+3.82)} &\bf \textcolor{themeblue}{(-0.51)} &\bf \textcolor{themeblue}{(+0.45)} \\
\midrule
\multirow{7}{*}{\makecell[l]{UCF101 \\ MiTv2 }}
&OpenMax  &\underline{86.33}&\underline{77.49}&\underline{44.40}&\underline{74.63} &93.29 &90.17 &29.84 & \underline{94.90}\\ 
&MC Dropout &76.61&62.32&48.43&74.24 &93.53 &90.97 &\underline{25.21} &94.11\\
&BNN SVI &83.13&76.20&48.63&73.15  &93.52 &91.24 &25.34 &93.89\\
&SoftMax &82.58&74.91&46.39&\underline{74.63} &92.62 &90.87 &30.55 &94.11\\
&RPL &81.47&73.98&49.62&71.89 &\underline{93.69} &\underline{92.04} &25.97 &94.26\\
&DEAR &81.48&77.03&77.58&74.42 &90.88 &90.55 &60.28 &93.89\\
&PSL(ours) &\bf{88.88}&\bf{83.30}&\bf{34.91}&\bf{78.69} &\bf{95.70} &\bf{95.06} &\bf{20.03} &\bf{95.51}\\
&$\mathbf{\Delta}$ &\bf \textcolor{themeblue}{(+2.55)} &\bf \textcolor{themeblue}{(+5.81)} &\bf \textcolor{themeblue}{(-9.49)} &\bf \textcolor{themeblue}{(+4.06)} &\bf \textcolor{themeblue}{(+2.01)} &\bf \textcolor{themeblue}{(+3.02)} &\bf \textcolor{themeblue}{(-5.18)} &\bf \textcolor{themeblue}{(+1.25)} \\
\bottomrule[1pt]
\end{tabular}
\vspace{-0.3cm}
\caption{OSAR performance under I3D backbone.}
\label{tab:bench_i3d}
\end{table*}

\begin{table*}[h!]
\centering
\tablestyle{3pt}{0.8}
\begin{tabular}{llcccccccc}
\toprule[1pt]
& & \multicolumn{4}{c}{\bf{w/o K400 Pretrain}} & \multicolumn{4}{c}{\bf{w/ K400 Pretrain}} \\ \cmidrule{3-10}
   \bf{Datasets} &\bf{Methods}                     & \bf{AUROC$\uparrow$}    & \bf{AUPR$\uparrow$}  &\bf{FPR95$\downarrow$}  & \bf{Acc.$\uparrow$}  & \bf{AUROC$\uparrow$}   & \bf{AUPR$\uparrow$} &\bf{FPR95$\downarrow$}   &  \bf{Acc.$\uparrow$}  \\ \midrule
  \multirow{7}{*}{\makecell[l]{UCF101 \\ HMDB51 }}
&OpenMax  &80.67	&
50.49	&52.46 &75.40 & 92.49  &  78.27 &  35.65   & 96.30\\ 
&MC Dropout &76.10	&41.37	&50.82 &75.16 & 91.83  &  77.71  &   29.82 &\underline{96.70}\\
&BNN SVI &\underline{81.66}	&\underline{56.72}	&49.66 &76.58 &  93.34  &  85.57  &   27.89   & 96.56\\
&SoftMax &79.15	&48.54	&\underline{48.79} &75.63	& \underline{93.82}  &  85.56  &  24.74   &       96.70\\
&RPL &81.35	&54.65	&51.64 &\underline{78.36} &93.81  &  85.41  &   \underline{24.06}   &       \bf{96.93}\\
&DEAR &78.00 &49.38	&68.49 &76.21	& 92.28  &  \underline{87.09}  &   62.99   &       96.48\\
&PSL(ours) &\bf{86.20}	&\bf{64.65}	&\bf{42.48} &\bf{79.40}	&\bf{95.24}  &  \bf{89.76}  &   \bf{18.72}   & 96.52\\
&$\mathbf{\Delta}$ &\bf \textcolor{themeblue}{(+4.54)} &\bf \textcolor{themeblue}{(+7.93)} &\bf \textcolor{themeblue}{(-6.31)} &\bf \textcolor{themeblue}{(+1.04)} &\bf \textcolor{themeblue}{(+1.42)} &\bf \textcolor{themeblue}{(+2.67)} &\bf \textcolor{themeblue}{(-5.34)} &\bf \textcolor{themeblue}{(-0.49)} \\
\midrule
\multirow{7}{*}{\makecell[l]{UCF101 \\ MiTv2 }}
&OpenMax  &79.60 &70.05	&51.08 &75.63 &94.34 &89.90 &25.42 &96.30\\ 
&MC Dropout &75.88	&63.12	&51.40 &75.63 &93.43  &  90.43  &   24.52   &       \underline{96.70}\\
&BNN SVI &\underline{82.89}	&
\underline{76.13}	&\underline{46.88} &76.58 &93.53  &  92.34  &   28.81   &       96.56\\
&SoftMax &51.08 &75.63	&79.60	&70.05	&94.67  &  93.34  &   22.14   &       96.70\\
&RPL &81.42	&73.07	&49.13 &\underline{78.36}&\underline{94.76}  &  \underline{93.39}  &   \underline{21.99}   &       \bf{96.93}\\
&DEAR &78.21&69.30	&62.02 &76.21&92.60  &  93.09  &   59.98   &       96.48\\
&PSL(ours) &\bf{85.00}&\bf{77.08}	&\bf{43.16}&\bf{79.40}&\bf{96.81}  &  \bf{96.22}  &   \bf{14.52}   &       96.52\\
&$\mathbf{\Delta}$ &\bf \textcolor{themeblue}{(+2.11)} &\bf \textcolor{themeblue}{(+0.95)} &\bf \textcolor{themeblue}{(-3.72)} &\bf \textcolor{themeblue}{(+1.04)} &\bf\textcolor{themeblue}{(+2.05)}
&\bf\textcolor{themeblue}{(+2.83)}
&\bf\textcolor{themeblue}{(-7.47)}
&\bf\textcolor{themeblue}{(-0.49)}\\
\bottomrule[1pt]
\end{tabular}
\vspace{-0.3cm}
\caption{OSAR performance under SlowFast backbone.}
\label{tab:bench_slf}
\vspace{0.5cm}
\end{table*}

\section{Representation analysis through singular value spectrum}

\begin{figure*}[h!]
\vspace{0.3cm}
  \centering
\includegraphics[width=0.99\linewidth]{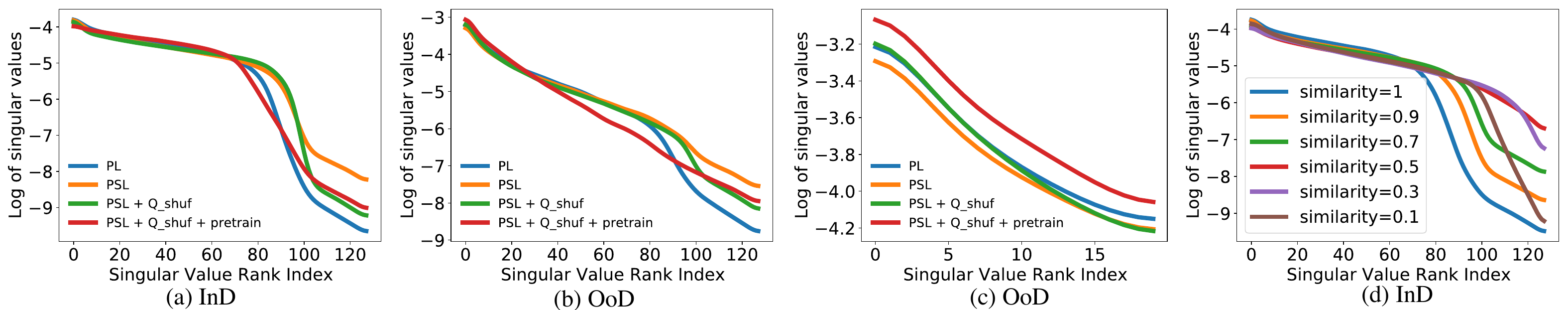}
  \vspace{-0.3cm}
  \caption{Singular value spectrum on HMDB51 (OoD) under different training conditions (a)-(c) and hyper-parameter $s$ (d). (c) contains the top 20 singular values in (b).}
  \label{fig:single}
  \vspace{0.5cm}
\end{figure*}

To deeply understand the feature representations learned by our method, we analyze the representation through singular value spectrum. We first compute the covariance matrix $C \in \mathbb{R}^{d \times d}$ of the embedding matrix:
\begin{equation}
    C= \frac{1}{M} \sum_{i=1}^{M}(z_{i}-\bar z)(z_{i}-\bar z)^{T},
\end{equation}
where $z_i$ and $\bar z_i$ denote the feature representation of a sample and mean representation of all samples respectively. $M$ is the total number of samples. Then we conduct singular value decomposition
on the matrix $C=USV^T, S=diag(\sigma^{k})$, and plot the singular values in sorted order and logarithmic scale $log(\sigma^{k})$. We provide the singular value spectrum in Fig.~\ref{fig:single}.
 
PSL has larger singular values than the PL in the larger rank index, illustrating that more information is contained in the not significant dimensions, which is reasonable as PSL keeps the IS information with no direct supervision signal, but these IS information does help for better OSAR performance according to Table 2 in the paper. PSL with shuffled samples $Q_{shuf}$ has larger singular values than PSL in the small rank index, indicating more diverse information is learned in the important dimensions, which are supposed to refer to CS information as CS information is learned by the explicit supervision signal. The closed-set accuracy with $Q_{shuf}$ is higher than without $Q_{shuf}$ in Table 2 further testifies our conclusion. In Tabel 2 we see that the representations of the same class are tighter with more CS information. Therefore, learning the distinct temporal information from shuffled videos can enlarge the open-set task related CS information while PSL can enlarge the IS information, which fulfills the goal to enlarge Eq. 3 for better OSAR performance.

\section{Open-set performance \textit{w.r.t.} $s$ with $Q_{shuf}$}

We provide extension results of Table 4 in the paper. The results are based on HMDB51 (OoD) from scratch. $s$ for $Q_{sc}$ is set as 0.7, and we change the value of $s$ for $Q_{shuf}$ in Table~\ref{tab:abla_s_q}. We can see that the performance is optimal when $s$ for $Q_{shuf}$ is 0.8, but the same $s$ with $Q_{sc}$ which is 0.7 also achieves the good performance. So to reduce the number of hyper-parameters, we pick up the same $s$ for $Q_{sc}$ and $Q_{shuf}$ by default. In addition, we can see that the closed-set accuracy is lower when $s=1$ compared to $s=0.8$. This is because we set the similarity between the original video and the shuffled video as 1, which is not reasonable as the temporal information is totally lost in the shuffled video.

\begin{table*}[h]
\centering
    \tablestyle{6pt}{1}
\begin{tabular}{ccccc}
\toprule[1pt]
 \bf{$s$} &\bf{AUROC$\uparrow$} & \bf{AUPR$\uparrow$} &\bf{FPR95$\downarrow$}  & \bf{Acc.$\uparrow$} \\ \midrule
1	&82.04	&53.82	&51.82 &72.89	\\
0.9 &83.12	&57.04	&46.84 &73.31\\
0.8 &\bf 86.43	&\bf 65.58	&\bf 41.75 &76.53\\
0.7 &85.25	&63.91	&48.34 &\bf 76.98\\
0.6 &85.26	&62.93	&46.89 &76.77\\
0.5 &84.08	&61.76	&53.53 &75.13\\
0.4 &82.75	&59.09	&52.72 &73.79\\
0.3	&77.34	&53.84	&68.14 &67.67\\
0.2 &73.94	&50.63	&75.55 &60.21\\
0.1 &68.86	&41.39	&82.15 &39.00\\
\bottomrule[1pt]
\end{tabular}
\vspace{-0.3cm}
\caption{Ablation results of different $s$ for $Q_{shuf}$.}
\label{tab:abla_s_q}
\end{table*}

\section{t-SNE visulization}

\begin{figure*}[h!]
  \centering
  \includegraphics[width=0.99\linewidth]{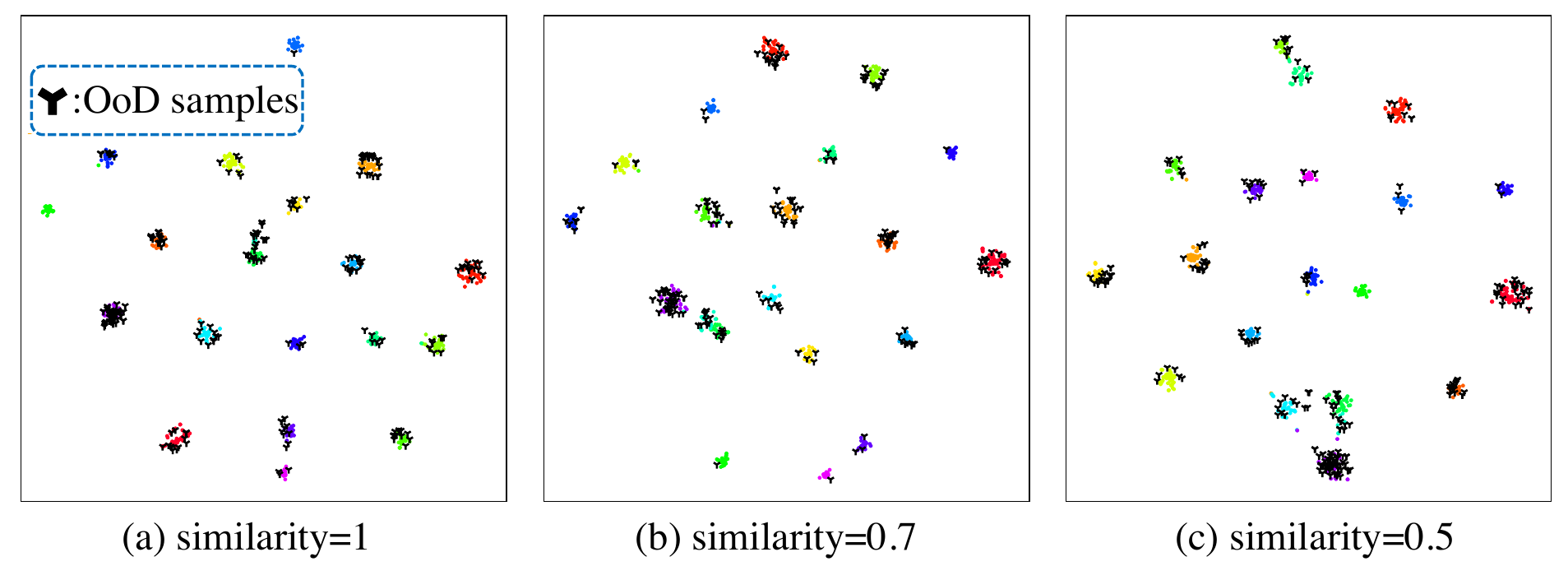}
  \vspace{-0.2cm}
  \caption{t-SNE visualization of PSL.}
  \label{fig:tsne_psl}
  \vspace{-0.2cm}
\end{figure*}

\begin{figure*}[h!]
  \centering
  \includegraphics[width=0.99\linewidth]{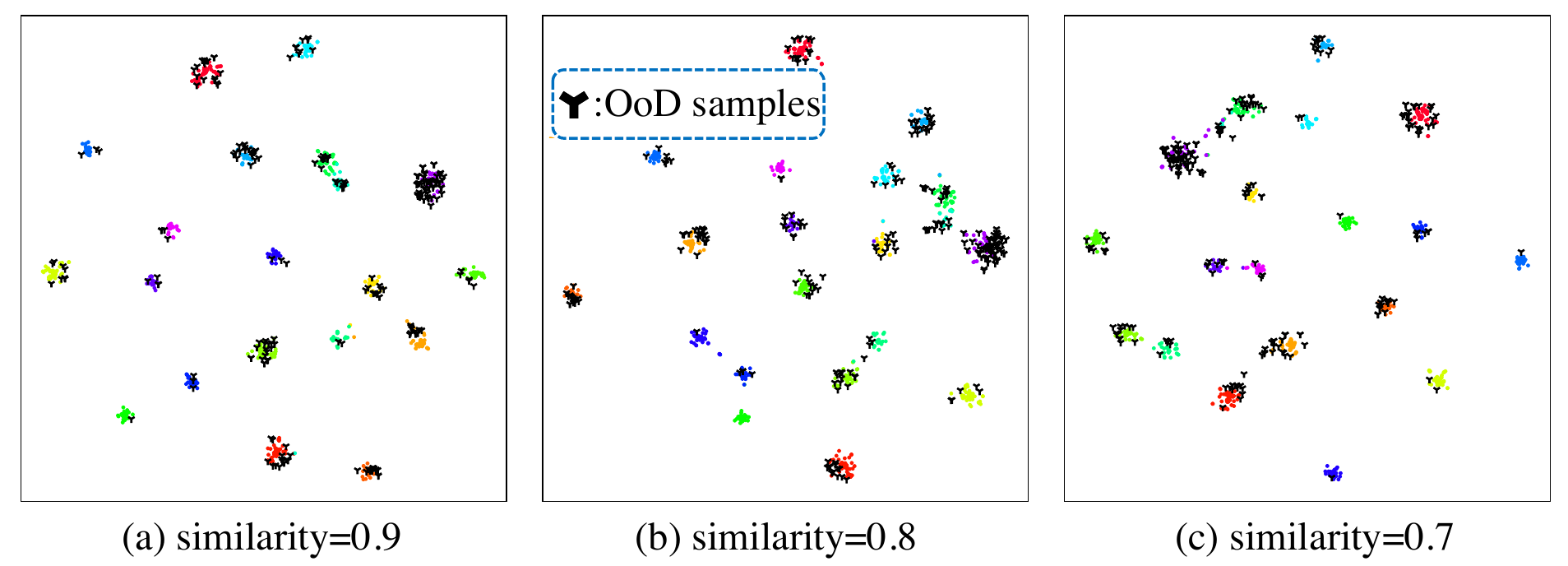}
  \vspace{-0.2cm}
  \caption{t-SNE visualization of PSL with $Q_{ns}$.}
  \label{fig:tsne_psl_n}
\end{figure*}

\begin{figure*}[h!]
  \centering
  \includegraphics[width=0.99\linewidth]{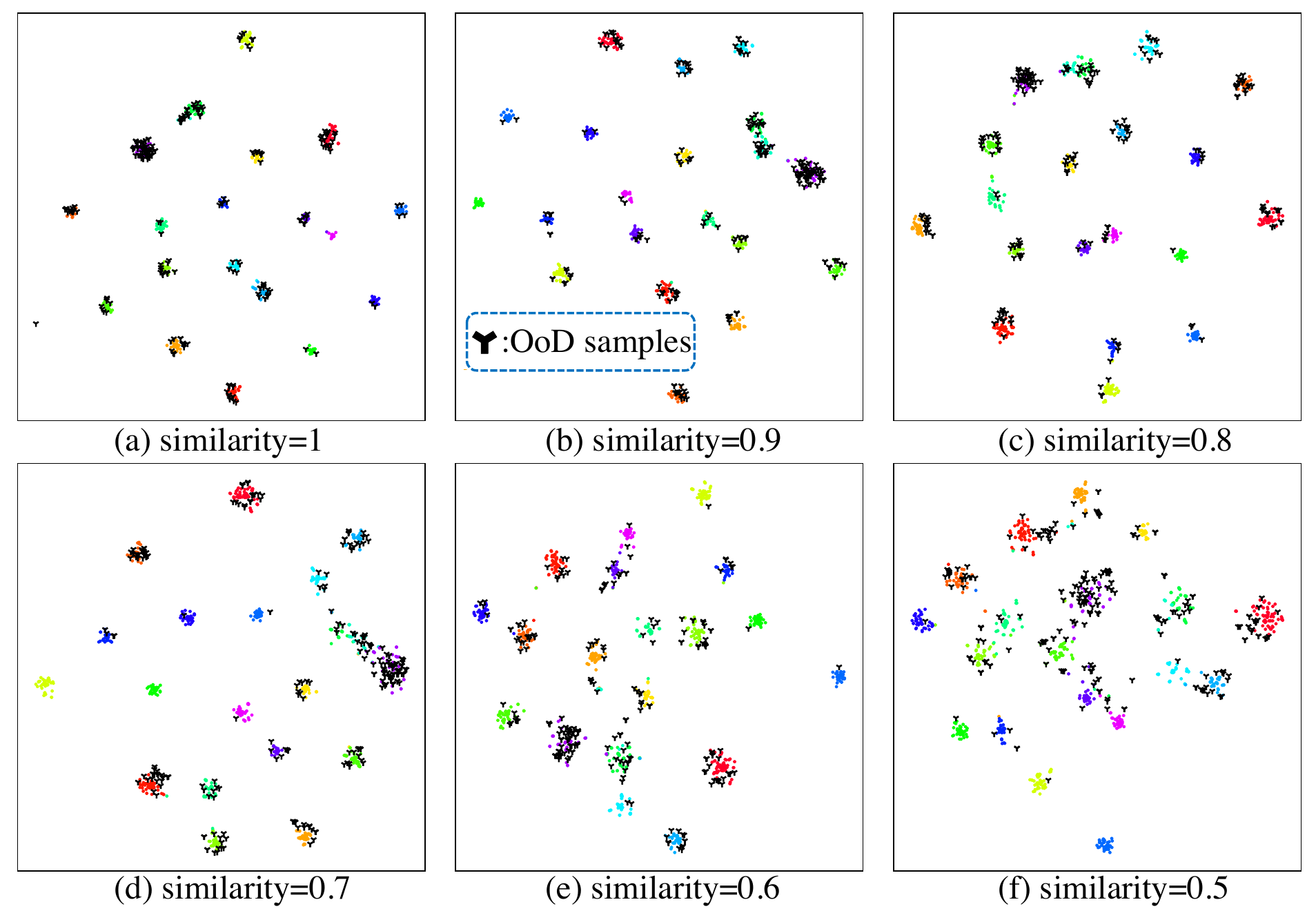}
  \vspace{-0.2cm}
  \caption{t-SNE visualization of PSL with $Q_{ns},Q_{sc}$.}
  \label{fig:tsne_psl_n_s}
\end{figure*}

\begin{figure*}[h!]
  \centering
  \includegraphics[width=0.99\linewidth]{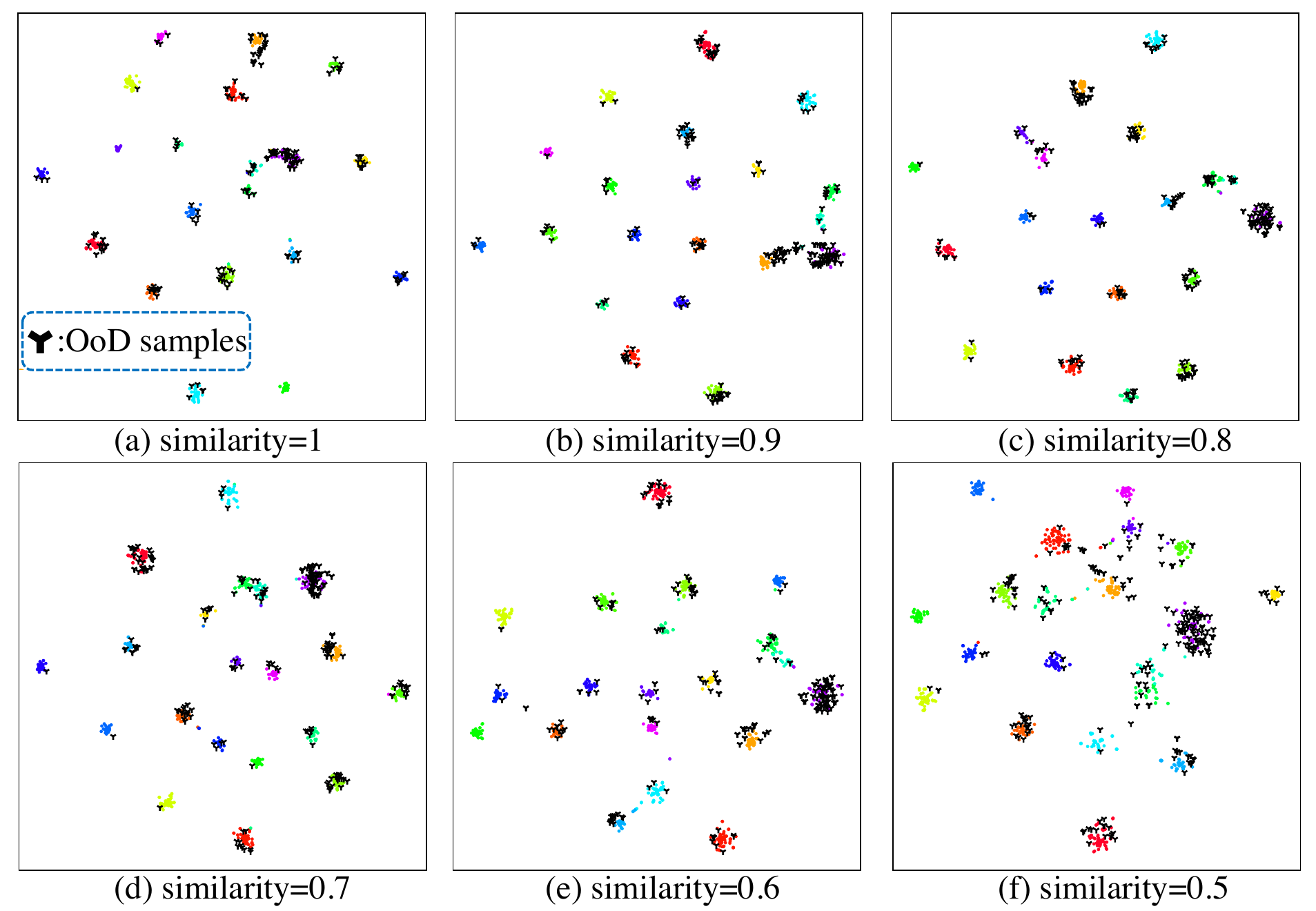}
  \vspace{-0.2cm}
  \caption{t-SNE visualization of PSL with $Q_{ns},Q_{sc},Q_{shuf}$.}
  \label{fig:tsne_psl_n_s_shu}
\end{figure*}

To illustrate the variance within a class, we provide the Table 2, Fig. 5 and 6 in the paper, which is enough to show the variance change due to different components in our PSL method. Here, we provide the t-SNE visualization for straight understanding. All results are based on HMDB (OoD) from scratch. We provide the visualization results of PSL, PSL with $Q_{ns}$, PSL with $Q_{ns},Q_{sc}$, and PSL with $Q_{ns},Q_{sc},Q_{shuf}$ in Fig.~\ref{fig:tsne_psl},~\ref{fig:tsne_psl_n},~\ref{fig:tsne_psl_n_s},~\ref{fig:tsne_psl_n_s_shu} respectively. From Fig.~\ref{fig:tsne_psl} we can see PSL alone cannot keep the intra-class variance when $s$ decreases. Fig.~\ref{fig:tsne_psl_n} and Fig.~\ref{fig:tsne_psl_n_s} tell us that $Q_{ns}$ and $Q_{sc}$ are important for PSL to keep the intra-class variance. Furthermore, $Q_{shuf}$ makes the feature representation tighter if we compare Fig.~\ref{fig:tsne_psl_n_s} and Fig.~\ref{fig:tsne_psl_n_s_shu}, which shows the model learns more CS information with $Q_{shuf}$.

\clearpage

\clearpage

\section{InD and OoD uncertainty distribution}

\begin{figure*}[t]
  \centering
  \includegraphics[width=0.95\linewidth]{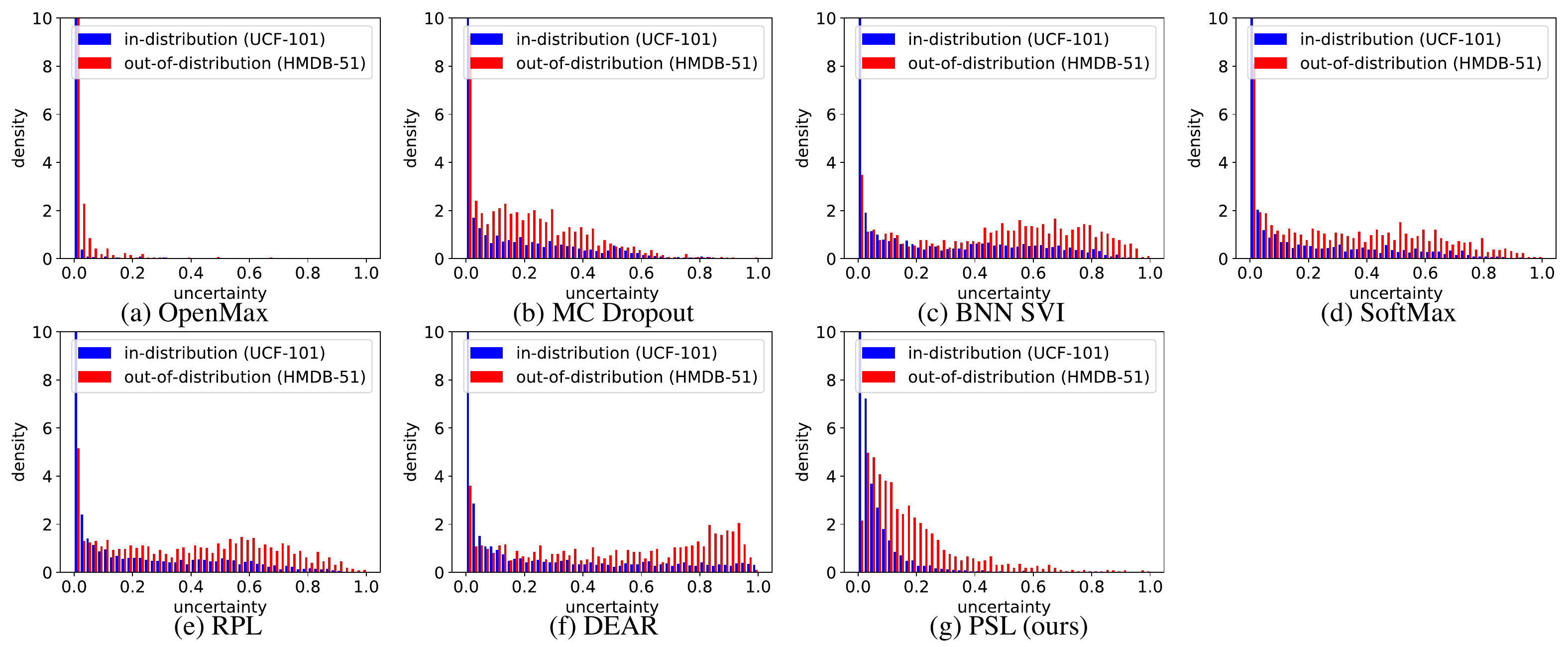}
  \vspace{-0.4cm}
  \caption{Uncertainty distribution on HMDB51 (OoD) w/o K400 pretrain.}
  \label{fig:dis_hmdb_sc}
  \vspace{-0.7cm} 
\end{figure*}

\begin{figure*}[h!]
  \centering
\includegraphics[width=0.95\linewidth]{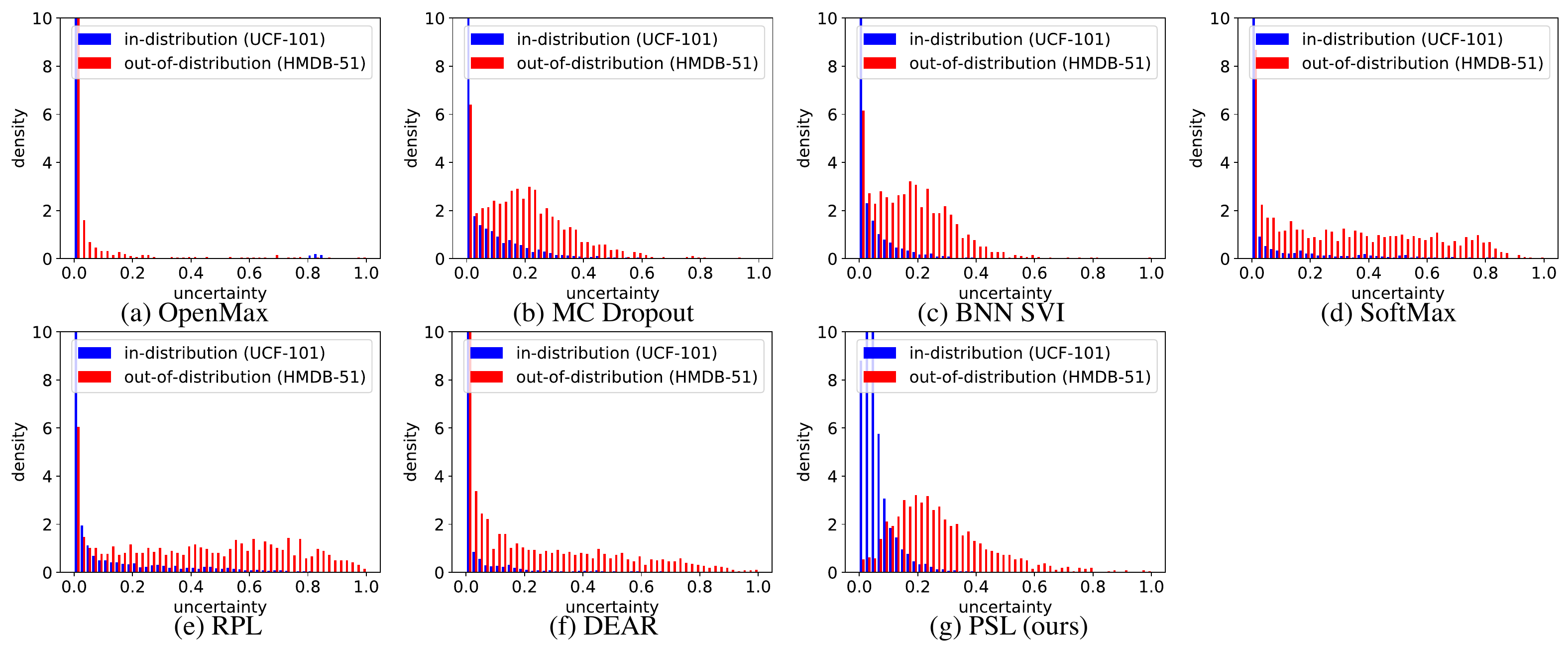}
  \vspace{-0.4cm}
  \caption{Uncertainty distribution on HMDB51 (OoD) w/ K400 pretrain.}
  \label{fig:dis_hmdb_ft}
  \vspace{-0.7cm}
\end{figure*}

\begin{figure*}[h!]
  \centering
  \includegraphics[width=0.95\linewidth]{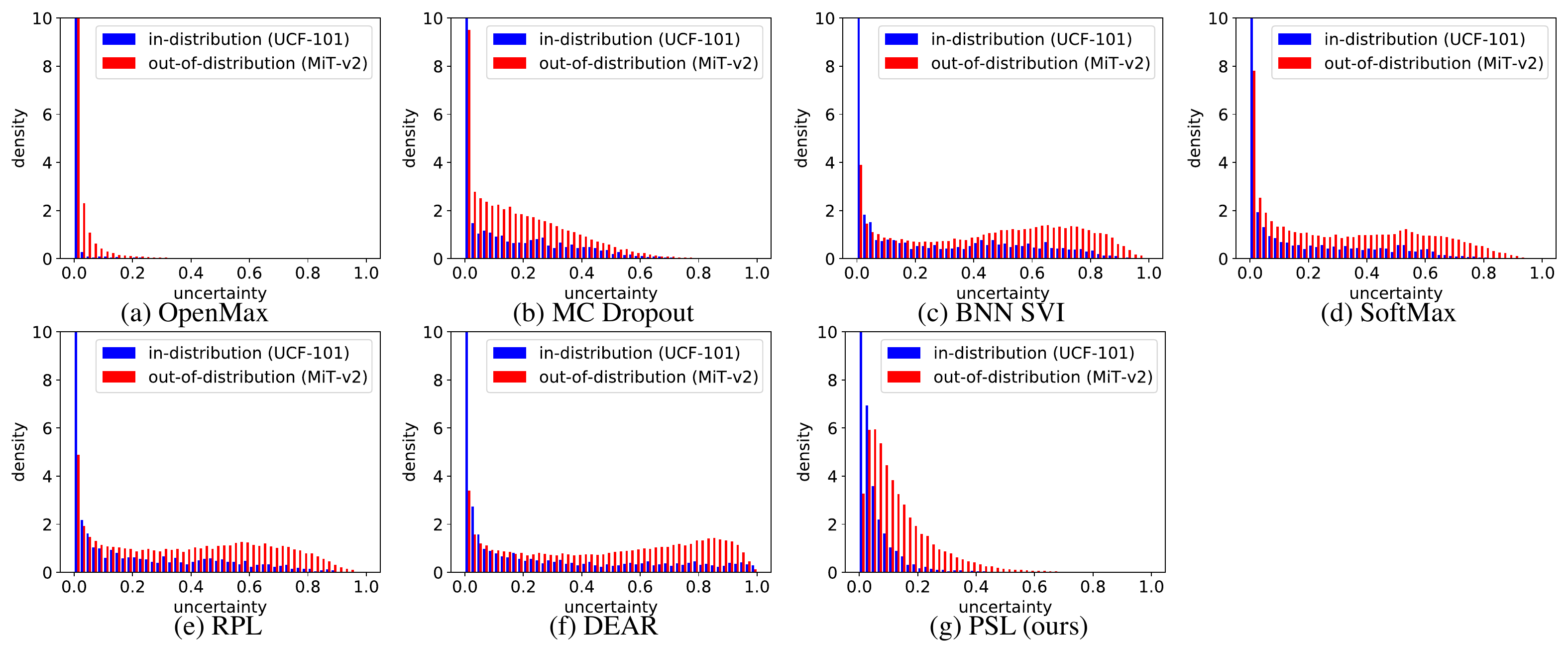}
  \vspace{-0.4cm}
  \caption{Uncertainty distribution on MiT-v2 (OoD) w/o K400 pretrain.}
  \label{fig:dis_mit_sc}
  \vspace{-0.2cm}
\end{figure*}

\begin{figure*}[t]
  \centering
  \includegraphics[width=0.95\linewidth]{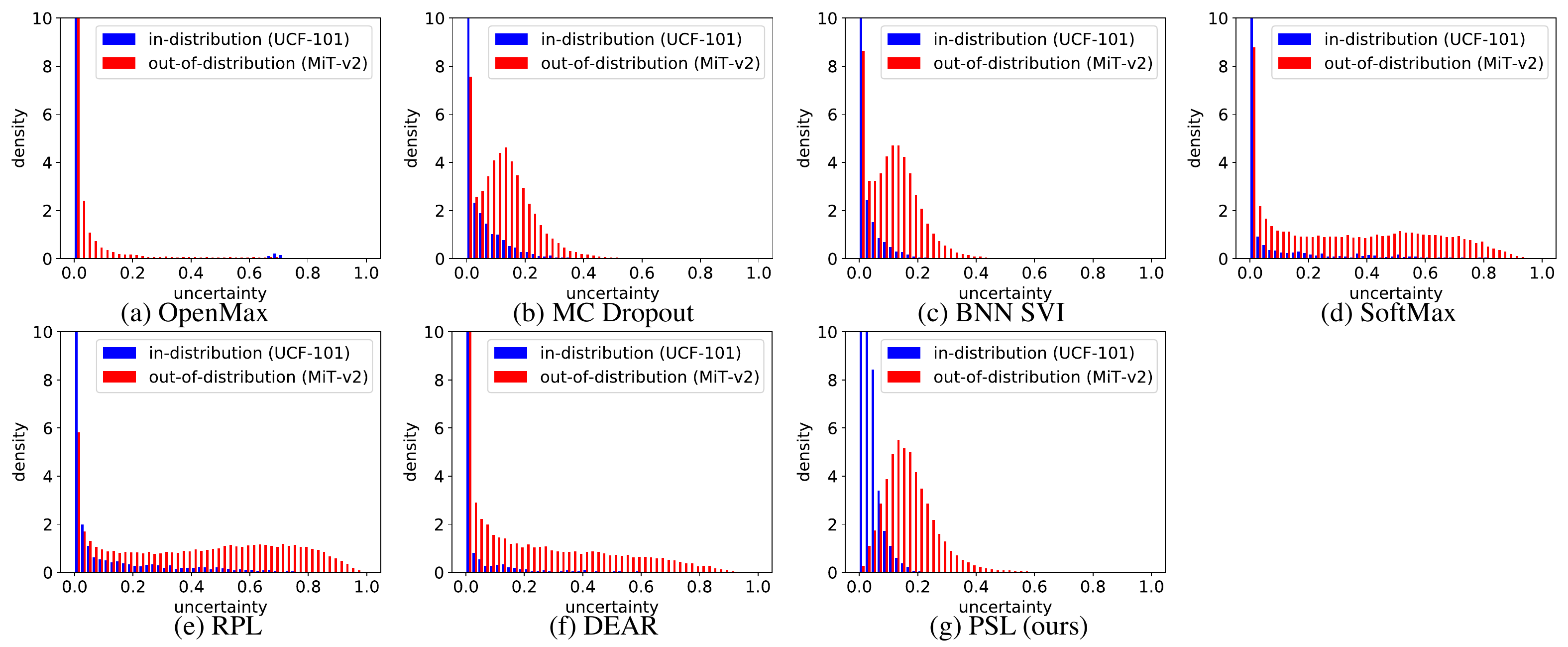}
  \vspace{-0.4cm}
  \caption{Uncertainty distribution on MiT-v2 (OoD) w/ K400 pretrain.}
  \label{fig:dis_mit_ft}
\end{figure*}

We provide the InD and OoD distribution on HMDB51 (OoD) and MiT-v2 (OoD) with K400 pretrain and without K400 pretrain. All results are based on TSM backbone for illustration. The results are shown in Fig.~\ref{fig:dis_hmdb_sc},~\ref{fig:dis_hmdb_ft},~\ref{fig:dis_mit_sc}, and~\ref{fig:dis_mit_ft}.

From Fig.~\ref{fig:dis_hmdb_sc} and~\ref{fig:dis_mit_sc} we can see that if there is no K400 pretrain, all methods have the overlapping uncertainty between InD and OoD distribution except OpenMax and our PSL. For instance, Fig.~\ref{fig:dis_hmdb_sc} (f) DEAR~\cite{bao2021evidential} shows the uncertainty of InD and OoD samples both cover the range from 0 to 1. In contrast, Fig.~\ref{fig:dis_hmdb_sc} (g) PSL shows that in our method, the InD distribution covers from 0 to 0.3, while the OoD distribution covers from 0 to 0.8. It means our method tends to assign higher uncertainty to OoD samples. For OpenMax, Fig.~\ref{fig:dis_hmdb_sc} (a) shows that InD uncertainy distribution is extremely close to 0, which is a good phenomenon, but the OoD uncertainty distribution only covers from 0 to 0.3, and the OoD samples whose uncertainty is larger than 0.3 is too sparse, which means OpenMax tends to assign low uncertainty to both InD and OoD samples, but assigner lower uncertainty to InD samples.

If we compare Fig.~\ref{fig:dis_hmdb_sc} to Fig.~\ref{fig:dis_hmdb_ft} or compare Fig.~\ref{fig:dis_mit_sc} to Fig.~\ref{fig:dis_mit_ft}, we can find that the InD distribution of all methods are closer to 0 with K400 pretrain. But all methods except our PSL have a serious over confidence problem, which is illustrated by the fact that the far left column of OoD samples is extremely high, which is also emphasized through the red circles in Fig. 4 of the paper. In contrast, the density of OoD distribution is highest at 0.2 uncertainty in our PSL method, and the density of OoD distribution is almost 0 at 0 uncertainty. Besides, it is very clear that the OoD distribution and InD distribution in our PSL is most distinguishable among all methods.
\clearpage
\twocolumn

{\small
\bibliographystyle{unsrt}
\bibliography{egbib}
}

\end{document}